\pdfoutput=1

\documentclass{article}
\usepackage[utf8]{inputenc}
\usepackage{authblk}
\usepackage{setspace}
\usepackage[margin=0.5in]{geometry}
\usepackage[pdftex]{graphicx}
\graphicspath{ {./figures/} }
\usepackage{subcaption}
\usepackage{amsmath}
\usepackage{makecell}
\usepackage{amssymb}
\usepackage{lineno}
\usepackage{hyperref}
\usepackage{threeparttable}
\usepackage{pdfpages}
\usepackage[pdftex]{graphicx}
\usepackage[edges]{forest}
\usepackage{tikz}
\usepackage{tikz-qtree}
\usepackage[figuresright]{rotating}
\usepackage{multirow}
\usepackage{color}
\usepackage{stackengine}
\usepackage{microtype}

\usepackage{xcolor}

\usepackage[style=nejm, 
citestyle=numeric-comp,
sorting=none]{biblatex}
\title{Artificial Intelligence for Biomedical Video Generation}

\author[1]{Linyuan Li}
\author[1]{Jianing Qiu}
\author[1]{Anujit Saha}
\author[2]{Lin Li}
\author[1]{Poyuan Li}
\author[1]{Mengxian He}
\author[1]{Ziyu Guo}
\author[1]{Wu Yuan}

\affil[1]{Department of Biomedical Engineering, The Chinese University of Hong Kong, Hong Kong SAR}
\affil[2]{Department of Informatics, King's College London, United Kingdom}
\affil[*]{Address correspondence to: wyuan@cuhk.edu.hk}

\date{}

\begin{document}
\begin{sloppypar}

\maketitle

\begin{abstract}
As a prominent subfield of Artificial Intelligence Generated Content (AIGC), video generation has achieved notable advancements in recent years. The introduction of Sora-alike models represents a pivotal breakthrough in video generation technologies, significantly enhancing the quality of synthesized videos. Particularly in the realm of biomedicine, video generation technology has shown immense potential such as medical concept explanation, disease simulation, and biomedical data augmentation. In this article, we thoroughly examine the latest developments in video generation models and explore their applications, challenges, and future opportunities in the biomedical sector. We have conducted an extensive review and compiled a comprehensive list of datasets from various sources to facilitate the development and evaluation of video generative models in biomedicine. Given the rapid progress in this field, we have also created a github repository to regularly update the advances of biomedical video generation at: \url{https://github.com/Lee728243228/Biomedical-Video-Generation}
\end{abstract}


\section{Introduction}

Artificial Intelligence Generated Content (AIGC) stands as a cornerstone in contemporary computer vision research, bolstering significant accomplishments across numerous sectors including economy~\cite{li2023Ecoagent}, healthcare~\cite{qiu2023large}, and transportation~\cite{zhang2023learning}. Notably within the biomedical field, image generation technology has been effectively applied to various imaging modalities, including Computed Tomography (CT)~\cite{mr-ct, diffusion-mri-ct}, Magnetic Resonance Imaging (MRI)~\cite{dai2020multimodal, jiang2023cola, qiu2023visionfm}, fundus photography~\cite{bellemo2019generative, go2024fundusgeneration}, and pathological imaging~\cite{moghadam2023morphology, yellapragada2024pathldm}. The fidelity of these generated images has seen progressive enhancements through technological evolution: from early generative models such as generative adversarial network (GAN)-based models~\cite{dai2020multimodal, bellemo2019generative}, to auto-regressive (AR) models~\cite{zhou2024conditional, vqgan}, then to cutting-edge diffusion models~\cite{diffusion-mri-ct, yellapragada2024pathldm}, and now moving towards the combination of auto-regressive and diffusion (AR + Diffusion) models~\cite{show-o, zhou2024transfusion}.

Compared to static images that contain solely spatial information, videos encompass additional temporal and motion features. For instance, cardiac ultrasound videos present not only the cardiac structure and pathological conditions but also the dynamics of cardiac valves and pumping functions. Hence, richer information content poses a more complex challenge for video generation. Thus far, predominant techniques in video generation are also based on GANs~\cite{saito2017temporal, li2018video, tulyakov2018mocogan}, AR models~\cite{kondratyuk2023videopoet, hong2022cogvideo}, and diffusion models~\cite{stable-video-diffusion, bar2024lumiere}. Following the robust generative capabilities demonstrated by Sora~\cite{sora} and Movie Gen~\cite{MovieGen}, diffusion models based on transformer architectures~\cite{diffusion-transformer, dubey2024llama3} have been increasingly employed for video generation, achieving state-of-the-art results. The AR + Diffusion models, by combining the strengths of AR models and diffusion models~\cite{zhou2024transfusion, show-o}, and featuring a more lightweight design~\cite{show-o}, have the potential to become the next-gen video generative models.

Despite the current video generation models' capability to simulate medical scenarios~\cite{li2024endora, zhao2024largeDSA, cho2024surgen, fundus2video, sun2024bora}, in order to achieve a more realistic and reliable simulation, considerations should be taken from three aspects: understanding principles in physics, effective evaluation metrics for generated medical content, and controllability and explainability of generation.

\textbf{Understanding principles in physics} is critical to enhancing the realism and precision of synthesizing biomedical videos. Taking surgical operations as an example for physics understanding, surgical procedures involve manipulating deformable tissues and organs using articulated instruments to achieve desired outcomes. While existing video generation models~\cite{li2024endora, sun2024bora} can create surgical scenes, they fail to model the surgical operations cohesively. The recent Movie Gen~\cite{MovieGen} has demonstrated certain abilities to simulate physical behaviors, but this ability has not been proven in the generation of medical videos. To achieve a more precise understanding of the motion characteristics in biomedicine, knowledge from physiology and pathology also needs to be further learned by generative models. 

In addition, it is crucial to understand the significance of~\textbf{evaluation criteria} for biomedical video synthesis. In addition to considering the coherence and authenticity of the generated content~\cite{ji2024t2vbench, fan2023aigcbench, t2v-compbench}, it is also necessary to ensure the medical utility and applicability of the generated content and its value added to the existing biomedical data. Therefore, in the design of evaluation criteria, it is necessary to additionally consider whether the biomedical knowledge contained in the generated content is meaningful and meets the needs of medical practice. 

Besides, generated videos could serve various medical purposes, such as aiding in diagnosis or education. Information to be generated in videos should be able to be precisely controlled, necessitating both \textbf{controllability and explainability}. Although controlling mechanisms such as proposed in ControlNet~\cite{controlNet} has proved its applicability in medical generation~\cite{guo2024maisi, sharma2024controlpolypnet}, many issues remain unresolved. 


Despite facing enormous challenges, the field of biomedical video generation is poised for exciting advancements. Addressing these challenges will open up numerous opportunities for technological innovation. With a foundation of existing video generation models and a wealth of biomedical video datasets, this article will explore potential application scenarios. By analyzing current technologies and challenges, we can pave the way for innovative developments in video generation that will greatly benefit the biomedical and healthcare communities.

\begin{figure}[htb]
    \includegraphics[page=1, width=1\columnwidth]{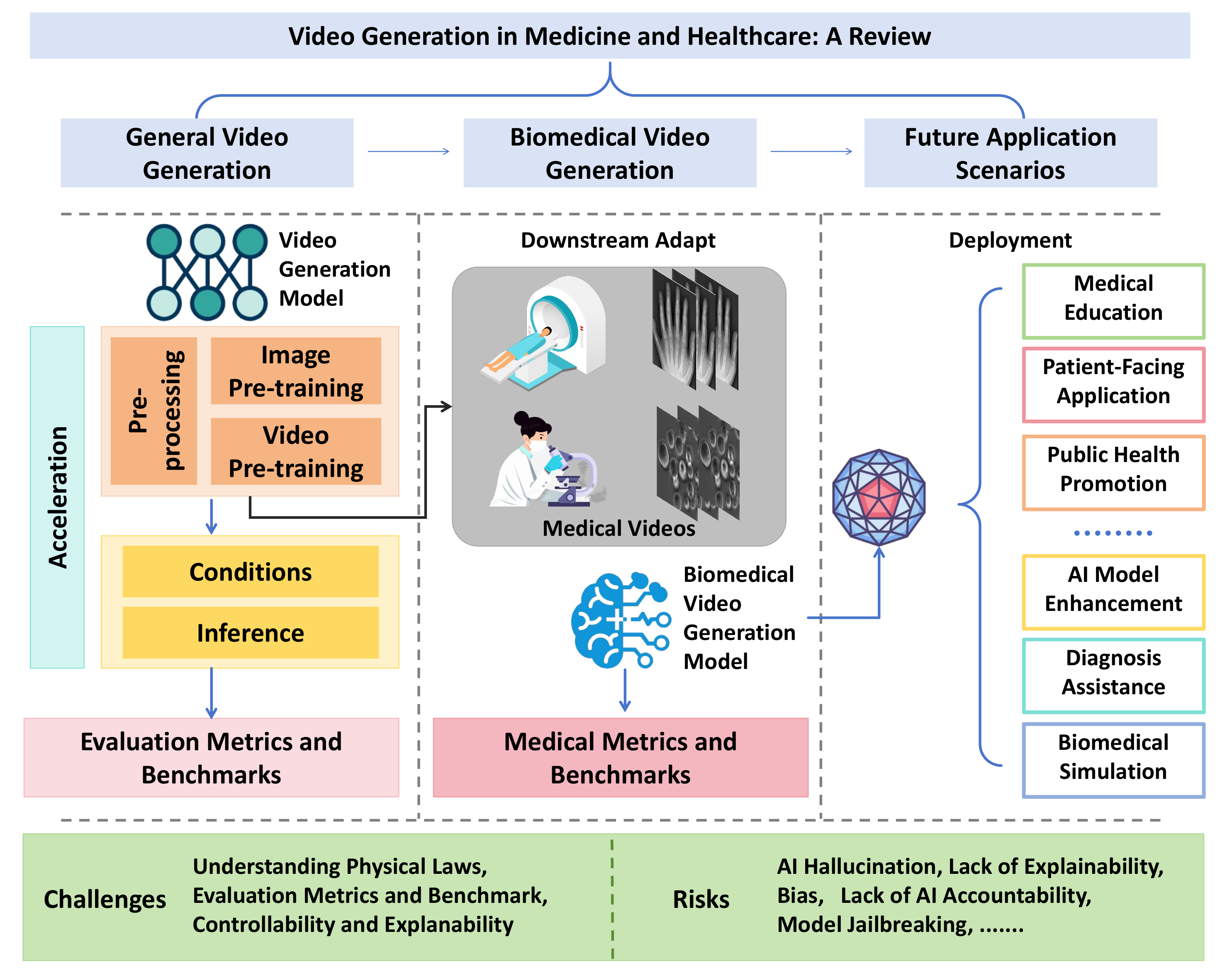}
    \caption{The existing workflow of biomedical video generation, including large-scale pretraining, adaptation in the biomedical domain, and deployment in biomedical scenarios. In the process of development and deployment of video generation technology, we highlight prominent challenges, such as learning medical physical laws, as well as risks related to hallucinations and bias.}
    \label{fig:ecosystem}
\end{figure}

The main contributions of this work are summarized
as follows: 
\begin{itemize}
    \item We summarized the three main challenges in medical video generation, including learning physical laws, establishing evaluation metrics and benchmarks, and enhancing controllability and explainability, and analyzed the corresponding potential solutions.
    \item We conducted a comprehensive investigation of existing video generation models in the general domain and surveyed models in biomedical and healthcare domains.
    \item We curated existing biomedical video generation datasets, including open-source datasets, video libraries, and biomedical videos on various multimedia platforms.
    \item We discussed the potential applications of video generation including medical education, patient-facing applications, and public health promotion, and analyzed their feasibility.
\end{itemize}

\textbf{Article Pipeline}: Section \ref{Challenges} analyzes the challenges faced by biomedical video generation and the potential solutions to address these challenges.
Section \ref{Workflow} introduces video generation workflow including data pre-processing, model architecture, training, inference, and evaluation. Section \ref{Medical Videos} focuses on medical video datasets and generation techniques applied in biomedicine. Section \ref{Application} discusses future directions of biomedical video generation and analyzes their feasibility and challenges. In section \ref{Risks}, we discuss noteworthy risks in biomedical video generation, and finally, we conclude in section \ref{Conclusion}. 

\section{Challenges}\label{Challenges}
 
The challenges in video generation primarily lie in modeling physical principles in biomedicine, devising effective and meaningful evaluation strategies, and enhancing generative controllability and explainability. This section reviews existing literature that sheds light on potential solutions to addressing these challenges in biomedical video generation.
\begin{figure}[htb]
    \includegraphics[page=1, width=1\columnwidth]{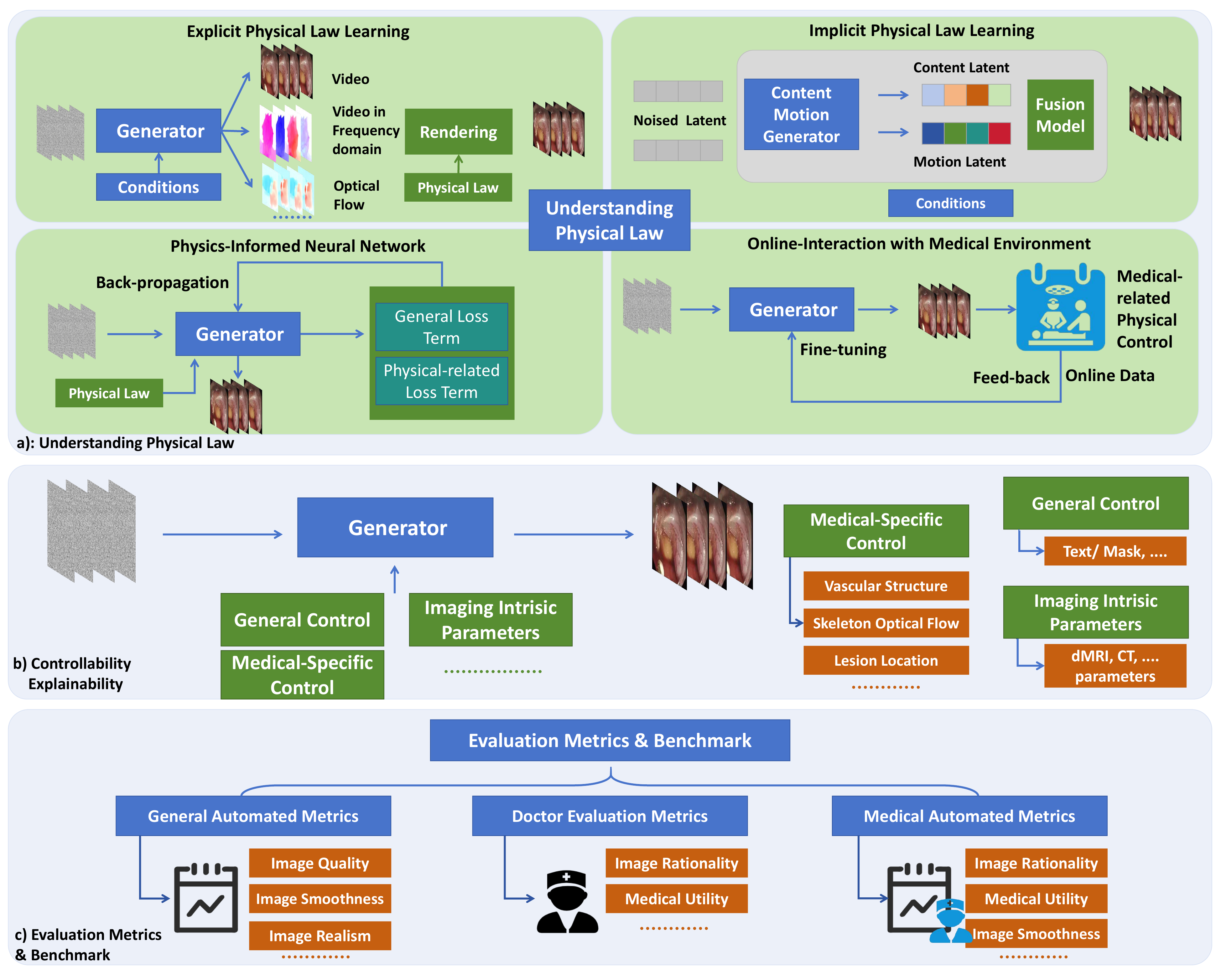}
    \caption{The challenges that biomedical video generation techniques face include a) understanding principles in biomedicine such as medical physics; b) controllability and explainability; and c) robust evaluation metrics and benchmarks.}
    \label{fig:ecosystem}
\end{figure}
\subsection{Understanding Physical Laws}\label{Physical Law Learning}

The understanding of physical laws indicates to the process of simulating and learning the physical phenomena such as the motion of objects, the effects of forces, and interactions between objects within the video content related to medicine. This process aims to ensure that the generated videos are not only visually realistic but also adhere to the principles of physical dynamics found in the real world. In addition to physical laws, other biomedical principles such as physiological and pathological laws are equally important.

Taking laparoscopic surgeries as an example, the physical phenomena observed are complex and multifaceted. Key aspects include: 1) Tissue Pulsation: The rhythmic pulsation of biological tissues, typically synchronized with the heartbeat, plays a critical role in understanding the physiological state of the patient during surgery. 2) Instrument Displacement: The movement of surgical instruments is crucial for performing precise manipulations. This displacement can be influenced by various factors, including the surgeon's technique and the properties of the tissues being operated on. 3) Deformation of Tissues: Surgical manipulation results in various types of deformation, including elastic deformation due to stretching and plastic deformation from cutting. These interactions can be complex and are influenced by the mechanical properties of both the instruments and the tissues. Through a comprehensive review of the literature, the modeling of physical laws in video generation can be categorized into four categories: 1)explicit physical law learning; 2)implicit physical law learning; 3)physics-informed neural networks; and 4) online interaction with the medical environment.

\textbf{Explicit Physical Law Learning} The methodologies for explicit physical law learning involve the direct synthesis of object motion using generative models, such as optical flow and frequency domain changes, followed by rendering on images or videos to achieve effective physical simulation~\cite{generative-image-dynamics, shi2024motion-I2V, zhang2024physdreamer, liu2024physgen}. For instance, GIT~\cite{generative-image-dynamics} employs diffusion models to learn information in the frequency domain and then integrates this information into one image for animation, thereby accomplishing the modeling of simple physical phenomena like vibrations.  PhysDreamer~\cite{zhang2024physdreamer} offers a more direct and interpretable approach to physical modeling by solving the differential equations of elastic materials using the Material Point Method (MPM)~\cite{xie2024physgaussian, jiang2016material} to model their physical laws. In the context of surgical scenarios, the relationship between stress and strain represents a primary form of physical deformation and the modeling of such a relationship can refer to PhyDreamer~\cite{zhang2024physdreamer}.

\textbf{Implicit Physical Law Learning} Implicit methods do not directly learn the physical phenomena, such as object motion trajectories or elastic deformation as explicit methods do~\cite{generative-image-dynamics, shi2024motion-I2V, liu2024physgen, zhang2024physdreamer}. Instead, they model object motion by learning motion features~\cite{cmd}. For example, generative models typically learn the temporal and spatial information of videos~\cite{stable-video-diffusion, videoLDM}, or the information related to motion and content separately~\cite{tulyakov2018mocogan, skorokhodov2022stylegan}. This decoupled approach is an implicit learning method; CMD~\cite{cmd} decomposes video content into content frames and motion latent representation during video encoding, which are then processed separately by a diffusion model. This decomposition method is frequently used when GANs are employed for video generation~\cite{tulyakov2018mocogan, skorokhodov2022stylegan}.

\textbf{Physics-Informed Neural Network} In traditional machine learning approaches, the learning process is primarily data-driven, with the model heavily reliant on large volumes of high-quality data. However, in practical applications, there is often a scarcity of data or the presence of noise, making it challenging for data-driven models to yield accurate and reliable results. To address this, incorporating physical knowledge as prior information aims to overcome the limitations of data insufficiency and make generation or prediction results more aligned with physical intuition~\cite{physics-informed-ml}. Particularly in the medical field, introducing the constraints of physical laws can make the model's decision outcomes more interpretable and controllable, thereby achieving a credible effect.

To incorporate physical laws into machine learning, these laws are typically embedded within the model's architecture~\cite{zhang2024phy-diff, tirindelli2021rethinking, momeni2021synthetic} and the design of the loss function~\cite{qiao2024zero}. In the visual generation domain, Phy-diff~\cite{zhang2024phy-diff} improves synthesized MRI quality by informing dMRI physical information (Diffusion coefficient map atlas) into the diffusion process with principled noise management, conditioned on the XTRACT atlas for anatomical details. Tirindelli et al.~\cite{tirindelli2021rethinking} augment ultrasound data by integrating physics-inspired transformations including deformations, reverberations, and
signal-to-noise ratio adjustments, aligned with the principles of ultrasound imaging, providing anatomically and physically consistent variability. Momeini et al.~\cite{momeni2021synthetic} model cerebral microbleeds with a Gaussian shape to simulate the data's characteristics including shape, intensity volume, and location, guided by MRI properties. In the generation domain, the incorporation of physical knowledge such as imaging properties helps boost the generation effect and performance of downstream tasks.

\textbf{Online Interaction with Medical Environment} Generated videos often include hallucinatory content and unrealistic physical effects. While expanding datasets~\cite{stable-video-diffusion, sora} and optimizing models~\cite{MovieGen, cmd} offer partial solutions, collecting external feedback from the environment is crucial and one effective solution for improving video generation~\cite{soni2024videoagent}.

By employing reinforcement learning, generative models are enabled to interact with the external environment, adjusting themselves based on a reward function to make the generated effects more consistent with the physical laws of the natural environment alleviating the issue of hallucinations to a certain extent.

VideoAgent~\cite{soni2024videoagent} is a representative of this approach. Based on the task description, it is necessary to capture a frame from the environment as the initial frame, and then the generative model generates a video based on the initial frame and task description. Subsequently, the content generated is judged against the standard of optical flow to determine if it complies with the physical laws. Content that conforms is collected and used to fine-tune the video generation model. When the model literally undergoes fine-tuning based on a variety of environments, it can then generate videos that adhere to the physical laws across different environments.
\subsection{Controllability and Explainability}\label{controllability and explainability}

Input conditions often contain rich medical information, which is a direct manifestation of medical utility. For example, masks may include information about the location and size of lesions, which provide critical information for physicians in making diagnoses. Therefore, enhancing the representation of control conditions in the generated content and ensuring the explainability of the generated content is important in medical generation. This section discusses the existing control methods, how to strengthen control, and the explainability of control, addressing the ways and difficulties in tackling these challenges.

\textbf{Control in Medical Domain} Text, mask, and depth information are common modalities used to control the generation of medical videos, which provide guidance to the generation process through control mechanisms such as ControlNet~\cite{li2024fairdiff, guo2024maisi, sharma2024controlpolypnet}, Clip~\cite{cho2024surgen, sun2024bora} and condition-specific encoder~\cite{zhou2024heartbeat, yu2024explainable} to the generation of medical images and videos. For instance, FairDiff~\cite{li2024fairdiff} generates point masks through a point-cloud diffusion model, which is then utilized to control the image diffusion model in generating fundus images. ControlPolypNet ~\cite{sharma2024controlpolypnet} controls the generation of polyp images based on segmentation masks.

\textbf{Controllability Enhancement} By integrating the inherent physical knowledge of medical imaging into the generation process, or by employing more rational and comprehensive modal control for generation, the effect of controllability can be enhanced. Phy-diff~\cite{zhang2024phy-diff} introduces inherent physical information of dMRI into the diffusion process and makes it more suitable for dMRI generation. This approach incorporates physical information to optimize the diffusion process also has the potential to be applied to the generation of images or videos from CT, fundus fluorescein angiography (FFA), ultrasound, and other modalities. Furthermore, in fundus2Video~\cite{fundus2video} work, a knowledge mask containing lesion and vessel information serves as an additional condition to assist in the generation of FFA sequences. HeartBeat~\cite{zhou2024heartbeat} utilizes a comprehensive set of modalities to guide ultrasound video generation in both coarse and fine-grained levels. Such conditions typically need to be determined by specific generation tasks and medical modalities.

\textbf{Explainability} In order to enhance the controllability of generative models, the explainability of the model has to be taken into account.
Although some existing works, such as ECM~\cite{yu2024explainable} and Heartbeat~\cite{zhou2024heartbeat}, can control and even explain the generation of medical videos, there are still issues to be addressed to enhance overall explainability, including but not limited to 1) the contribution of different modalities to the content of the generated video needs to be measured quantitatively; 2) whether there is redundancy between modalities; and 3) whether excessive control modalities can lead to the degradation of model performance.

\subsection{Evaluation Metrics and Benchmarks}\label{challenges for metrics}

The introduction of benchmarks like vbench~\cite{ji2024t2vbench} has allowed for a more comprehensive evaluation of video generation, which employs $16$ different metrics to thoroughly assess video generation models. Given the unique characteristics of medical videos, a comprehensive benchmark is essential and in particular, with specialized metrics tailored for medical videos.

\textbf{General Automatic Metrics} General automatic metrics refer to assessment standards that are applicable to both natural video generation models and medical video generation models. Commonly used evaluation metrics include Fréchet Inception Distance (FID)~\cite{fid}, Inception Score (IS)~\cite{IS}, Fréchet Video Distance (FVD)~\cite{FVD}, Structural Similarity Index (SSIM)~\cite{ssim}, Peak Signal-to-Noise Ratio (PSNR), and CLIPSIM~\cite{clip}, among others. These metrics could assess multiple dimensions of the quality of generated videos, such as realism, smoothness of imagery, and the alignment of conditions with generation outcomes. However, there is a domain gap between medical videos and natural videos, hence the evaluation metrics for natural video generation models cannot be directly applied to medical video generation models without consideration of their discrepancy.

\textbf{Doctor Evaluation Metrics} Doctor evaluation refers to the direct involvement of a physician in assessing the synthesized videos, and providing scores based on their individual preferences. Due to the absence of benchmarks designed for medical image or video generation, this study refers to an ophthalmology report generation benchmark known as 
FFA-IR~\cite{li2021ffa} that not only takes into account the general evaluation metrics for report generation, such as Recall-Oriented Understudy for Gisting Evaluation(ROUGE) and Bilingual Evaluation Understudy(BLEU) but also incorporates the assessment by ophthalmologists of the generated reports. The assessment criteria include the fluency of the report, the rationality of the lesion description within the report, and the accuracy of the described lesion location, among other professional medical issues. Similarly, in video generation, it is also necessary to consider whether the generated content adheres to physiological laws and medical utilities with the involvement of medical professionals.

\textbf{Medical Automatic Metrics} Physicians' participation in evaluating synthesized videos is typically time-consuming and inefficient. Hence, automatically integrating medical expertise into evaluation metrics is a promising alternative. On one hand, using LLMs as substitutes for physicians in evaluations is an option.
Medial expert LLMs~\cite{luo2022biogpt} and multimodal LLMs~\cite{li2024llavaMed} possess the potential to accomplish this task. On the other hand, designing medical evaluation metrics is also feasible. Since there are no such evaluation metrics in the field of medical visual generation, evaluation metrics in radiologist report generation can be taken as references such as F1-RadGraph~\cite{jain2021radgraph}, CheXBert vector similarity~\cite{yu2023evaluating}, and RadCliQ~\cite{yu2023evaluating} used in MultiMedEval~\cite{royer2024multimedeval}. RadGraph constructs a radiology knowledge graph based on reports and then calculates the similarity between the generated report and the reference report's graphs to obtain a more specialized evaluation metric. When assessing medical video generation models, this unique relationship can also be captured, such as building a graph model that uses detection and classification techniques to model the relationships between surgical instruments and biological tissues, achieving a specialized evaluation of surgical video generation.

\section{Technical Background of Video Generation}\label{Workflow}
This section outlines the essential technical aspects of video generation, including data preparation and pre-processing, the design of neural network architectures, the training and inference of generative models, as well as evaluation metrics and benchmarks.

\subsection{Data Preparation and Pre-processing}\label{pre-process}
Video generative models require extensive video data to develop. Recent surveys~\cite{zhu2024sora-simulator,xing2023survey,cho2024sora-worldmodel} have conducted a detailed investigation of the datasets, highlighting commonly employed datasets in general domains such as UFC101~\cite{soomro2012ucf101}, WebViD~\cite{bain2021frozen}, and LAION-5b~\cite{schuhmann2022laion}. These datasets, however, are mainly for text-guided video generation and unconditional video generation. Specifically, text-guided video generation aims to generate videos that represent text semantics, and the unconditional video generation model synthesizes videos by sampling noises from Gaussian Distribution, which represents the distribution of training data. For video generation conditioned on other modalities, such as depth-guided or pose-guided video generation, the conditional modalities can either be manually curated or synthesized using specialized tools like a depth-generator~\cite{midas} and pose-generator~\cite{controlNet, guler2018densepose}.

After gathering the datasets for training, the first step is to pre-process the videos along with their corresponding conditional modalities. For unconditional generation, only videos have to be pre-processed. As the most common conditional modality is text, the discussion below will focus on video-text data pre-processing.

\begin{itemize}
    \item \textbf{Video Pre-processing} Videos have to be pre-processed to facilitate training. Common strategies include cut detection, filtering static scenes, resizing, and downsampling. SVD~\cite{stable-video-diffusion} introduces a series of comprehensive and effective video pre-processing methods, including cascaded cut detection by PySceneDetect~\cite{PySceneDetect}, Keyframe-aware clipping, and calculating optical flow-based motion scores, aesthetic scores, clip scores~\cite{clip}, and OCR detection rates for clip filtering. These processes remove clips that are motionless, unclear, unaesthetic, and mismatched with the text, as well as text-video pairs with low-quality text.

    \item \textbf{Text Pre-processing} Text diversity and richness are important for text-video generation~\cite{somepalli2023understanding}. In addition to text resources from datasets, they can also be obtained from a captioning model~\cite{zhou2024streaming, chen2023surgical, EILEV} and video subtitles. Text enhancement is commonly used to obtain more detailed and diverse descriptions. LLMs~\cite{achiam2023gpt, brown2020gpt-3, touvron2023llama, luo2022biogpt} and multimodal LLMs~\cite{li2023videochat, maaz2023videochatgpt, li2024llavaMed} can be used to create more diverse descriptions for videos. 
    
\end{itemize}

\subsection{Design of Neural Network Architecture}\label{Model-Architecture}

\begin{figure}[htb]
    \includegraphics[page=1, width=1\columnwidth]{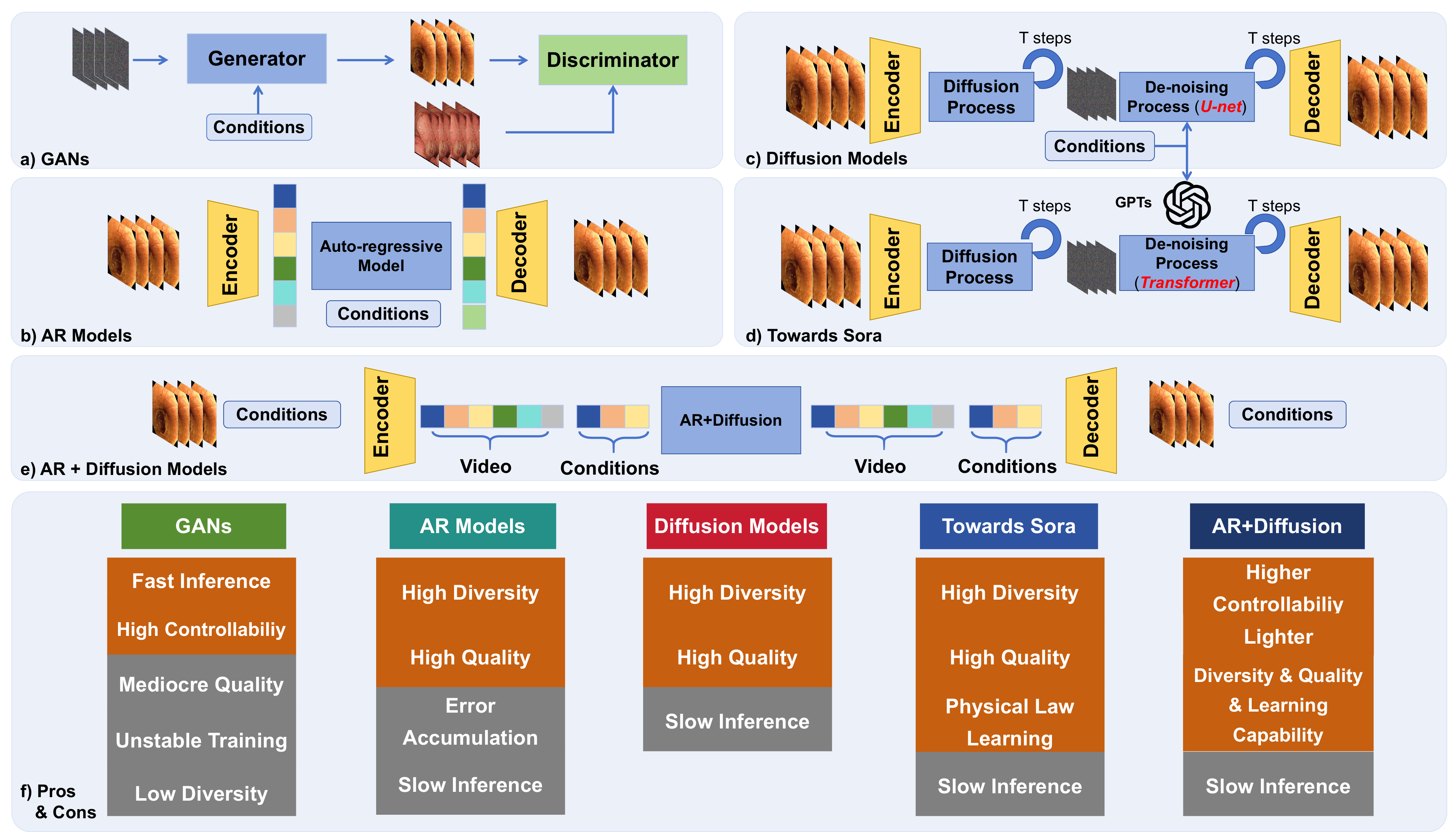}
    \caption{Architectures of video generation models, and their strengths and limitations. GANs, AR models, diffusion models, and Sora-alike models have been widely used in video generation. The AR + Diffusion models, due to their capabilities in understanding and generating content, holds promise for reliable biomedical video generation.
}
    \label{fig:Architecture}
\end{figure}

The architectural landscape of video generation models has evolved rapidly over the past few years. Early approaches mainly adopted variational autoencoder (VAE)~\cite{wu2021godiva, walker2021predicting} and flow-based~\cite{kingma2018glow, kumar2019videoflow} video generation techniques. Later, studies predominantly utilized GANs~\cite{GAN, tulyakov2018mocogan, saito2017temporal}, followed by the rising popularity of AR models for video generation~\cite{wu2021godiva, huang2023factor}. Recently, research on diffusion models~\cite{he2022latent, stable-video-diffusion} has significantly advanced video generation technology. The new generation of diffusion models, exemplified by Sora~\cite{sora} and Movie Gen~\cite{MovieGen}, features a more diverse training dataset, a more efficient compression model, and a more powerful diffusion backbone. The breakthrough in the diffusion backbone also lays the foundation for subsequent work~\cite{ma2024latte, chen2024gentron, henschel2024streamingt2v, bahmani2024vd3d, sora, MovieGen}. The AR + Diffusion models~\cite{zhou2024transfusion, show-o} demonstrate strong capabilities in image-text understanding and generation, showing potential in the field of medical visual generation. This section will primarily focus on analyzing GANs, AR video generation models, diffusion-based video generation models, Sora-alike Transformer-based video generation models, and AR+Diffusion models that show potential in medical video generation.

\textbf{GANs}~\cite{GAN, brock2018large, karras2019style, perarnau2016invertible, PD-GAN} comprise two principal components: a generator and a discriminator. The generator synthesizes outputs by sampling a noise $z$ from prior distribution $p_g(z)$, denoting $G(z)$, and the discriminator, functioning as a classifier, aims to discern whether the output is real or generated by the generator. In video generation, GANs often employ a decoupled approach; for instance, VGAN~\cite{vgan} generates the foreground and background streams using 3D and 2D convolutional networks respectively, which are subsequently amalgamated to produce a cohesive video. Furthermore, GAN framework frequently utilizes decoupling strategies along various dimensions, such as spatial dimensions and temporal dimensions~\cite{saito2017temporal}, as well as content dimensions and motion dimensions~\cite{skorokhodov2022stylegan,yu2022generating}, to enhance the generation process.

GAN models generate videos in only one step with the generator, showing its high efficiency. Controllability is commendable because the generation conditions can directly control the latent features, rather than controlling the latent noise as diffusion models do~\cite{draggan}. However, in terms of generation quality, GANs typically do not match the performance of large-scale models such as diffusion models and auto-regressive models~\cite{diffusionbeatgan}. Due to the inherent structure of GAN models, mode collapse can occur, which also leads to generation diversity not being particularly competitive~\cite{srivastava2017veegan}.

\textbf{AR Models} have been widely used in image generation~\cite{generative-from-pixel, lee2022autoregressive, yu2021diverse, yu2022scaling} based on the Vision Transformer~\cite{vision-transformer}. In auto-regressive generation, each new token is produced based on preceding tokens, a principle that has been extended to video generation as well.
For example, CogVideo~\cite{hong2022cogvideo} encodes video frames into a latent sequence and is trained auto-regressively, where the next frame is generated conditioned on the previous frames. VideoPoet~\cite{kondratyuk2023videopoet} applies an LLM~\cite{brown2020gpt-3} as its backbone and auto-regressively trains the input text, audio, and video tokens.

AR models are capable of generating high-quality and diverse visual content~\cite{dalle-3}. However, despite the model's ability to generate high-quality and diverse samples, the generative performance of these models will decline due to the drawback of error accumulation~\cite{lee2023mathematical}, 
as AR models generate subsequent tokens based on previous ones. In addition, these models are usually large-scale, so optimizing their training and inference speed is a challenge.

\textbf{Diffusion Models}~\cite{nichol2021improved, NEURIPS2020_4c5bcfec}, inspired by non-equilibrium thermodynamics~\cite{SohlDickstein2015DeepUL}, begin by defining a Markov chain that incrementally adds Gaussian noise to the data $x_0$, a process known as diffusion. Given the properties of the Markov process, the state of $x_t$ only depends on $x_{t-1}$. Subsequently, the model learns to reverse this diffusion process to reconstruct the desired data sample by removing the noise. In a diffusion model framework, the inputs are initially condensed using compression models such as VAE~\cite{vae} and VQ-VAE~\cite{vq-vae}, and followed by the Denoising Diffusion Probabilistic Model (DDPM) process~\cite{NEURIPS2020_4c5bcfec} guided by conditions such as text prompt~\cite{dreambooth, blip-diffusion, galimage, ge2023expressive}, images~\cite{ge2023expressive, xu2024prompt, mo2024freecontrol, voynov2023sketch}, audio\cite{yang2023align,qin2023gluegen}, or other modalities~\cite{chen2023seeing, 10205187, sun2024conformal}, and finally reconstructed by a decoder. Furthermore, diffusion models incorporate mechanisms that decouple video components, typically separating videos into temporal and spatial dimensions~\cite{videoLDM}. This approach enhances the model's ability to learn video features more effectively.

Diffusion-based generative models have become the mainstream approach for video generation due to their high generation quality and richness of details. However, because of its multi-step inference during generation, their training and inference speeds are relatively slow. Additionally, compared to the latent features in GANs, the noise latent poses a challenge in terms of explainability~\cite{draggan}.

\textbf{Towards Sora} Sora~\cite{sora} is an advanced text-to-video generation model developed by OpenAI that is capable of generating up to one minute of high-quality video content based on user input. The demonstration of Sora has accelerated the trend of using Transformer~\cite{transformer} as backbones for diffusion models. Sora, according to OpenAI's technique report~\cite{sora-report} and open-source projects~\cite{opensora, opensora-plan}, comprises three key components akin to Latent Diffusion Model(LDM)~\cite{latent-diffusion-model}, including a compression model~\cite{vae, vq-vae}, a conditioning mechanism, and a generation model. Diffusion Transformer (DiT)~\cite{dit} is the core of Sora, where noised latent from the diffusion process is patchified into a series of tokens and then added with positional embeddings. DiT blocks utilize adaLN~\cite{perez2018film} as the conditioner to introduce timestep condition $t$ and class label condition $c$. Notably, Sora makes adaLN initialized to zero to expedite large-scale training in a supervised learning context~\cite{goyal2017accurate}, referred to as adaLN-zero.

Movie Gen~\cite{MovieGen} is a media generator developed by Meta that integrates both video and audio generation capabilities. Movie Gen has designed a diffusion backbone based on the transformer architecture, which structurally draws inspiration from the design of LLama 3~\cite{dubey2024llama3}, and trains in flow matching manner~\cite{lipman2022flow}. However, it does not incorporate structures such as causal attention and Grouped-Query Attention (GQA) that are utilized for the auto-regressive training of LLama 3. It uses a temporal autoencoder (TAE) as its compression model combining VAE~\cite{vae} and a 1D convolution layer. During TAE training, it adds outlier penalty loss (OPL) to eliminate the negative impact of `spot' artifacts.

\textbf{AR + Diffusion Models} Auto-regressive models such as LLMs~\cite{achiam2023gpt, brown2020gpt-3, touvron2023llama, dubey2024llama3} and multimodal LLMs~\cite{wang2024longllava, liu2024llava} have demonstrated strong comprehension capabilities for text and multimodal information, while diffusion models like LDM~\cite{latent-diffusion-model, stable-diffusion} have shown impressive visual generation abilities. Transfusion~\cite{zhou2024transfusion} and Show-O~\cite{show-o} effectively integrate these two capabilities, exhibiting understanding and generation capabilities that are comparable to LLMs and diffusion models, along with enhanced text-image alignment capabilities.

Transfusion~\cite{zhou2024transfusion} utilizes a transformer as a generative model, employing causal attention when training text tokens and bidirectional attention when training image generation models. It integrates the next prediction loss $L_{LM}$ from LLMs and the $L_{DDPM}$ from diffusion models as its training objectives. Show-O~\cite{show-o} is consistent with Transfusion in terms of language training, but in image training, it adopts a mask token prediction loss $L_{MTP}$ to reconstruct the masked image patches. Comparatively, both models possess the capabilities of LLM diffusion models in text generation and image generation, and they have also achieved a new SOTA in text-video alignment. Show-O performs slightly better than Transfusion in terms of alignment. Although such models have not yet been applied to the generation of medical visual content, their demonstrated stronger condition-alignment capabilities can provide better explainability and controllability for medical content generation, holding potential in the generation of medical images and videos.

\subsection{Training}
\subsubsection{Workflow}

A general training strategy for developing video generation models involves large-scale pre-training on natural images, then large-scale pre-training on natural videos, followed by fine-tuning with high-quality videos, and finally domain-specific adaptations to downstream scenarios, e.g., medical video generation.

\textbf{Large-scale Image Pre-training} Image pre-training serves as the initial phase in training video generation models, with the primary goal of enriching the models' comprehension of visual content representations. Images are typically pre-trained on image generation models, but some models, such as Movie Gen~\cite{MovieGen}, treat images as a single-frame video and complete image pre-training on a video generation model.

\textbf{Large-scale Video Pre-training} Video pre-training enables the model to learn motion representation based on the model pre-trained on images with rich visual priors

Datasets such as ShareGPT4Video~\cite{sharegpt4video}, FineVideo~\cite{FineVideo}, WebVid10M~\cite{webvid100m}, and Kinetics-600~\cite{kinetics-600} offer rich resources for video training. ShareGPT4Video includes millions of videos featuring wild animals, cooking, sports, landscapes, and beyond, coupled with abundant textual information. However, low-quality data can degrade model performance. Therefore, video data and its modalities must undergo a series of pre-processing and filtering steps (see Section~\ref{pre-process}) to further enhance model performance.

Following image pre-training, the generative model undergoes further training. However, there are two key differences: 1) It is necessary to introduce a temporal dimension or motion features into the image generation model to model dynamic content, such as Make-A-Video~\cite{singer2022make} and LVDM~\cite{yang2023align}, which incorporate temporal attention, VDM~\cite{ho2022video} and Make-An-Animation~\cite{azadi2023make}, which introduce pseudo 3D covolution, and StyleGan-V~\cite{skorokhodov2022stylegan}, which incorporates motion vectors; 2) This step will introduce various conditions, such as text as a supervisory signal to guide video generation, promoting the model's learning of the consistency between conditions and generated content.

\textbf{High-Quality Video Fine-tuning} Following the previous two steps, training on higher-quality datasets can yield enhanced and more realistic outcomes. 

High-quality video datasets are normally high-resolution. For example, LSMDC~\cite{movie-description} provides a video dataset with 1080p resolution, while DAVIS~\cite{davis} includes videos with a resolution of 1280p, and RDS~\cite{videoLDM} offers a substantial collection of video clips with a dimension of $1024\times512$ pixels.

At this stage, it is essential to design fine-tuning tasks that help the model gain a deeper understanding of video content and the consistency between video conditions. Models such as the diffusion model and the auto-regressive model, e.g., SVD~\cite{stable-video-diffusion} and VideoPoet~\cite{kondratyuk2023videopoet}, have designed a series of self-supervised learning tasks to assist the model in further exploring the intrinsic features of videos. These tasks include video interpolation, video prediction, and image-to-video generation tasks.

\textbf{Downstream Adaptation} High-quality fine-tuning establishes a robust foundational model for subsequent downstream tasks. A notable application involves the transfer of generative models to the medical domain for medical video synthesis. A suite of generative tasks can be accomplished based on existing medical video datasets (see Section~\ref{Medical Videos} and Table~\ref{tab:medical video dataset}).

\subsubsection{Training Acceleration}
Video generation models often face issues such as slow training speeds and excessive consumption of resources like memory during training. Therefore, it is necessary to optimize the training process of these models to enhance training efficiency.

\textbf{GAN Acceleration} The training process of GANs is often characterized by instability due to its structural nature. Techniques such as Earth Mover's Distance~\cite{wgan}, Gradient Penalty~\cite{wgan-gp}, and TTUR~\cite{TTUR} help stabilize GAN training. Besides, skip-layer excitation and self-supervised discriminator~\cite{liu2020towards} reduce the number of parameters, thereby facilitating faster and more stable training.

\textbf{AR Model Acceleration} Acceleration techniques for auto-regressive models often align with those developed for transformers, especially LLMs, as most AR generators utilize LLM structure as their generation backbones~\cite{vqgan,kondratyuk2023videopoet, zhou2024conditional}. Models can be distributed across multiple GPUs for efficient training such as DeepSpeed~\cite{rasley2020deepspeed} and Zero~\cite{rajbhandari2020zero}. A series of parameter-efficient learning methods can also be applied including Lora~\cite{lora, qlora}, Adapter~\cite{adapt-tuning}, BitFit~\cite{bitfit}, and P-tuning~\cite{p-tuning}. Other efficient acceleration methods include model quantization~\cite{liu2023llm}, and FlashAttention~\cite{dao2022flashattention}.

\textbf{Diffusion Model Acceleration} Diffusion models often entail numerous steps for image/video generation, emphasizing the importance of optimizing this process for efficiency. SpeeD~\cite{speeD} separates time steps into distinct areas and accelerates training process. P2~\cite{choi2022p2} and Min-SNR~\cite{hang2023min_snr} adjust weights on the time steps based on heuristic rules. State Space Models(SSM) is utilized in DiffuSSM~\cite{diffusion-wo-att} to increase training speed without attention mechanism.

\subsection{Inference}

\subsubsection{Workflow}

During the training phase, the specified condition serves as a directive for the generation, thereby guiding the creation process. These conditions include modalities such as text, images, depth information, and sound (see Table~\ref{videoinference}). Beyond directly inputting conditions for the model to generate video content, there are various inference techniques that can boost the inference performance. These include but are not limited to: 1) prompt engineering; 2) video interpolation and super-resolution; 3) self-condition consistency model; and 4) explicit generation.

\textbf{Prompt Engineering} Since existing video generation models are trained with long and detail-rich prompts~\cite{yang2024cogvideox, sora, MovieGen}, a good prompt directly affects the quality of video generation, such as the richness of details. Prompt engineering is the technique that helps unleash their potential. For example, Sora~\cite{sora} utilizes GPT~\cite{brown2020gpt-3} to turn users' short prompts into longer captions to generate high-quality videos that better align with users' input. 

\textbf{Video Interpolation and Super-resolution} Videos directly generated by generative models may sometimes have a low frame rate and low pixel resolution. Therefore, post-processing with video interpolation models and super-resolution models can produce higher-quality videos. Make-A-Video~\cite{singer2022make}, ImagenVideo~\cite{ho2022imagen}, and VideoLDM~\cite{videoLDM} utilize a multi-stage process including a series of interpolation models and super-resolution videos to generate longer and higher-resolution videos.

\textbf{Vision-Language Model-guided Inference} VideoAgent~\cite{soni2024videoagent} proposed Vision-Language Model (VLM)-guided video generation. VideoAgent first plans the video conditioned on the first frame and language. Based on the video plan and latent noise from the previous iteration, VLM helps refine the model by making denoising adjustments. Besides VLM, humans can also engage in the refining iteration to adjust based on their own preferences. VLM-guided inference can effectively eliminate hallucinations and make the imagery more realistic.

\textbf{Explicit Generation} Explicit generation, proposed by EMUVideo~\cite{girdhar2023emu} divides the video generation process into two stages: 1) image generation conditioned on the text and 2) video generation conditioned on text and generated image. The image generated after the first stage is regarded as the explicit representation of generation content. It initializes the video generation model with stable diffusion model~\cite{stable-diffusion} weights and fine-tune the temporal layers on video datasets. This explicit inference framework helps generate videos with more details.

\begin{table}[h!]
\scriptsize
  \begin{center}
    \caption{Current video generation methods categorized by conditional modalities.}
    \label{videoinference}
    \begin{tabular}{c|c|c|c|c|c} 
      \textbf{Method} & \textbf{Year} &\textbf{Architecture}&\textbf{Method} & \textbf{Year} &\textbf{Architecture}\\
      \hline
      \multicolumn{6}{l}{\scriptsize  Text-guided Generation }\\
      \hline
       TGANs-C\cite{pan2017create}&2017&GAN&T2V\cite{li2018video}&2018&GAN\\
       
       StoryGAN\cite{li2019storygan}&2019&GAN&LVT\cite{latent-video-transformer}&2020&Auto-regressive Model\\
       Godiva\cite{wu2021godiva}&2021&VAE&LVDM\cite{he2022latent}&2022&Diffusion Model(\textcolor{blue}{U-net})\\
       
       VDM\cite{ho2022video}&2022&Diffusion Model(\textcolor{blue}{U-net})&ImagenVideo\cite{ho2022imagen}&2022&Diffusion Model(\textcolor{blue}{U-net})\\
       
       Stylegan-V\cite{skorokhodov2022stylegan}&2022&GAN&CogVideo\cite{hong2022cogvideo}&2022&Auto-regressiveModel\\
       
       Make-a-Video\cite{singer2022make}&2022&Diffusion Model(\textcolor{blue}{U-net})&Show-1\cite{zhang2023show}&2023&Diffusion Model(\textcolor{blue}{U-net})\\
       
       SVD\cite{stable-video-diffusion}&2023&Diffusion Model(\textcolor{blue}{U-net})&Stream2V\cite{henschel2024streamingt2v}&2024&Diffusion Model(\textcolor{blue}{U-net})\\

       Latte\cite{ma2024latte}&2024&Diffusion Model(\textcolor{red}{Transformer})&CogVideoX\cite{yang2024cogvideox}&2024&Diffusion Model(\textcolor{red}{Transformer})\\
       
       Gentron\cite{chen2024gentron}&2024&Diffusion Model(\textcolor{red}{Transformer})&VD3D\cite{bahmani2024vd3d}&2024&Diffusion Model(\textcolor{red}{Transformer})\\

       Sora\cite{sora}&2024&Diffusion Model(\textcolor{red}{Transformer})&MovieGen\cite{MovieGen}&2024&Diffusion Model(\textcolor{red}{Transformer})\\
       
       \hline
       \multicolumn{6}{l}{\scriptsize Pose-guided Generation}\\
       \hline
 
        DynamicGAN\cite{natarajan2022dynamic}&2022&GAN&DreamPose\cite{karras2023dreampose}&2023&Diffusion Model(\textcolor{blue}{U-net})\\

       MagicAnimate\cite{xu2024magicanimate}&2024&Diffusion Model(\textcolor{blue}{U-net})&MimicMotion\cite{zhang2024mimicmotion}&2024&Diffusion Model(\textcolor{blue}{U-net})\\

       Follow-your-pose\cite{ma2024follow}&2024&Diffusion Model(\textcolor{blue}{U-net})&Disco\cite{wang2024disco}&2024&Diffusion Model(\textcolor{blue}{U-net})\\

       \hline
       \multicolumn{6}{l}{\scriptsize Motion-guided Generation}\\
       \hline

       DragNUWA\cite{yin2023dragnuwa}&2023&Diffusion model(\textcolor{blue}{U-net})&MCDiff\cite{chen2023motion}&2023&Diffusion Model(\textcolor{blue}{U-net})\\

       DreamVideo\cite{wei2024dreamvideo}&2024&Diffusion Model(\textcolor{blue}{U-net})&VMC\cite{jeong2024vmc}&2024&Diffusion Model(\textcolor{blue}{U-net})\\

       MotionClone\cite{ling2024motionclone}&2024&Diffusion Model(\textcolor{blue}{U-net})&MotionCTRL\cite{wang2024motionctrl}&2024&Diffusion Model(\textcolor{blue}{U-net})\\

       Revideo\cite{mou2024revideo}&2024&Diffusion Model(\textcolor{blue}{U-net})&360DVD\cite{wang2024360dvd}&2024&Diffusion Model(\textcolor{blue}{U-net})\\

       \hline
       \multicolumn{6}{l}{\scriptsize Image-guided Generation}\\
       \hline
       Imaginator\cite{wang2020imaginator}&2020&GAN&LaMD\cite{hu2023lamd}&2023&Diffusion Model(\textcolor{blue}{U-net})\\
        
       GID\cite{generative-image-dynamics}&2023&Diffusion Model(\textcolor{blue}{U-net})&LFDM\cite{ni2023conditional}&2023&Diffusion Model(\textcolor{blue}{U-net})\\

       \hline
       \multicolumn{6}{l}{\scriptsize Depth-guided Generation}\\
       \hline

       Animate-a-Stroy\cite{he2023animate}&2023&Diffusion Model(\textcolor{blue}{U-net})&Make-your-video\cite{xing2024make}&2024&Diffusion Model(\textcolor{blue}{U-net})\\

       \hline
       \multicolumn{6}{l}{\scriptsize Sound-guided Generation}\\
       \hline

       TPOS\cite{jeong2023power}&2023&Diffusion Model(\textcolor{blue}{U-net})&Aadiff\cite{lee2023aadiff}&2023&Diffusion Model(\textcolor{blue}{U-net})\\

       Generative Disco\cite{liu2023generative}&2023&Diffusion Model(\textcolor{blue}{U-net})&TA2V\cite{zhao2024ta2v}&2024&Auto-regressive Model\\
       
       \hline
       \multicolumn{6}{l}{\scriptsize Video-guided Generation}\\
       \hline
       \multicolumn{6}{l}{Video Editing}\\
       \hline
       Videop2p\cite{video-p2p}&2023&Diffusion Model(\textcolor{blue}{U-net}) &Dreamix\cite{dreamix}&2023&Diffusion Model(\textcolor{blue}{U-net})\\

       DynVideo\cite{dynvideo}&2023&Diffusion Model(\textcolor{blue}{U-net})&Anyv2v\cite{anyv2v}&2023&Diffusion Model(\textcolor{blue}{U-net}) \\

       MagicCrop\cite{magicprop}&2023&Diffusion Model(\textcolor{blue}{U-net})&ControlAVideo\cite{controlavideo}&2023&Diffusion Model(\textcolor{blue}{U-net}) \\

       &CCedit\cite{ccedit}&2024&Diffusion Model(\textcolor{blue}{U-net})\\
       \hline
       \multicolumn{6}{l}{Video Interpolation \& Video Prediction}\\
       \hline
       PhBI\cite{meyer2015phase}&2015&Phase-based&AdaConv\cite{niklaus2017video}&2017&kernel-based\\

       SRFI\cite{deng2019self}&2017&kernel-based&PhaseNet\cite{meyer2018phasenet}&2018&Phase-based\\

       CyclicGen\cite{liu2019deep}&2019&OpticalFlow&VQI\cite{xu2019vqi}&2019&Optical Flow\\

       FIGAN\cite{FIGAN}& 2022&GAN&VIDIM\cite{jain2024videointerpolation}&2024&Diffusion Model(\textcolor{blue}{U-net})\\

       LDMVFI\cite{danier2024ldmvfi}&2024&Diffusion Model(\textcolor{blue}{U-net})\\

       \hline
       \multicolumn{6}{l}{\scriptsize Brain-guided Generation}\\
       \hline

       f-CVGAN\cite{wang2022reconstructing}&2022&GAN&CinimaticMindscapes\cite{chen2024cinematic}&2024&Diffusion Model(\textcolor{blue}{U-net})\\

       DynamicVStimuli\cite{yeung2024neural}&2024&Diffusion Model(\textcolor{blue}{U-net})&Animate-your-thoughts\cite{lu2024animate}&2024&Diffusion Model(\textcolor{blue}{U-net})\\

       \hline
       \multicolumn{6}{l}{\scriptsize Multi-modal-guided Generation}\\
       \hline

       NUWA\cite{wu2022nuwa}&2022&VAE&VideoPoet\cite{kondratyuk2023videopoet}&2023&Auto-regressive Model\\
       MovieFactory\cite{zhu2023moviefactory}&2023&Diffusion Model(\textcolor{blue}{U-net})&MovieComposer\cite{wang2024videocomposer}&2024&Diffusion Model(\textcolor{blue}{U-net})\\

       Lumieire\cite{bar2024lumiere}&2024&Diffusion Model(\textcolor{blue}{U-net})&Sora\cite{sora}&2024&Diffusion Model(\textcolor{red}{Transformer})\\

       AV-DiT\cite{wang2024av-dit}&2024&Diffusion Model(\textcolor{red}{Transformer})\\

       \hline
       \multicolumn{6}{l}{\scriptsize Unconditional Generation}\\
       \hline
       VGAN\cite{vgan}&2016&GAN&WGAN\cite{saito2017temporal}&2017&GAN\\
       WGAN\cite{arjovsky2017wasserstein}&2017&GAN&MocoGAN\cite{tulyakov2018mocogan}&2018&GAN\\

       DVD-GAN\cite{clark2019adversarial}&2019&GAN&DIGAN\cite{yu2022generating}&2022&GAN\\
       
    \end{tabular}
  \end{center}
\end{table}

\subsubsection{Inference Acceleration}

Inference speed and memory resource utilization are also crucial metrics for generative models. To address these challenges, a range of optimization algorithms have been developed. Optimization efforts primarily target auto-regressive models and diffusion models.

\textbf{AR Model Acceleration} Currently, the predominant architecture for mainstream auto-regressive generative models is LLM structure, aligning the acceleration methods of LLMs accordingly. DeepSpeed~\cite{rasley2020deepspeed} used in training acceleration, can also be utilized in inference acceleration. In addition, packages like vLLM~\cite{vLLM}, lightLLM~\cite{Light-LLM}, and TensorRT~\cite{TensorRT} have optimized cache management, attention, and quantization mechanism, which can significantly accelerate the inference of LLMs.

\textbf{Diffusion Model Acceleration} Parallel inference techniques offer a promising avenue for expediting diffusion-based generation. ParaDiGMS~\cite{shih2024parallel} and DistriFusion~\cite{li2024distrifusion} splits denoising process and input patches to different GPUs respectively to accelerate generation to reduce the computational cost. Time steps~\cite{xue2024accelerating, time-tuner} can also be optimized during inference time to increase efficiency. In addition, DeepCache~\cite{ma2024deepcache} and AT-EDM~\cite{AT-EDM} demonstrate the advantageous utilization of model cache and input data to enhance inference performance. To facilitate acceleration, specialized libraries such as Xformers~\cite{xformer}, AITemplate~\cite{AITemplate}, TensorRT~\cite{TensorRT}, and OneFlow~\cite{yuan2021oneflow} can be leveraged for streamlining the inference process.

\subsection{Evaluation Metrics and Benchmarks}\label{Natural Video Metrics}
\begin{sidewaystable}
\scriptsize
  \begin{center}
    \caption{Commonly used metrics for assessing the quality of generated videos.}
    \begin{tabular}{c|c|c|c|c} 
      \textbf{Metrics} & \textbf{Formula} &\textbf{Explanation}&\textbf{Function}&\textbf{Drawback}\\
      \hline
      \multicolumn{4}{l}{General Synthetic Video Metrics}\\
      \hline
        IS\cite{IS}& $\textit{IS}_{video} = exp(E(\frac{1}{T}\sum\limits_{t=1}^{T}E\left[KL(p(y_t|x_t)||p(y_t))\right])) $&\makecell{$IS$ measures $KL$ divergence \\between class probabilistic distribution \\of one generated frame $p(y_t|x_t)$ \\and distribution of all \\generated images $p(y)$, where \\$T$ is the number of frames }&\makecell[c]{Evaluation of diversity of \\generated samples and degree \\of one sample belongs to a \\certain class}&\makecell[c]{Only consider generated \\samples' distribution.}\\
        
        FID\cite{fid}& $d^2((m, C), (m_w, C_w)) = ||m - m_w||_2^2 + Tr(C + C_w - 2(CC_w)^{\frac{1}{2}})$&\makecell{$FID$ measures Fr\'echet distribution distance \\between ground truth and \\generated samples \\whose mean and co-variance \\are $(m, C)$ and $(m_w, C_w)$ respectively.}&\makecell[c]{Evaluation of Gaussian \\distribution distance between \\generated and real data \\in the feature level.}&\makecell[c]{Unable to assess the \\overfitting situation of \\generative models. Gaussian \\distribution is insufficient \\to represent feature \\distribution.}\\
        
        FVD\cite{FVD}& $MMD^2(q, p) = \sum\limits_{i \neq j}^m\frac{k(x_i, x_j)}{m(m-1)} - 2 \sum\limits_{i=1}^m\sum\limits_{j=1}^n\frac{k(x_i, y_j)}{mn} + \sum\limits_{i\neq j}^n\frac{k(y_i, y_j)}{n(n-1)}$&\makecell{$FVD$ measures distance \\between ground truth $p(X)$ \\and generated samples $q(Y)$ \\by Maximum Mean Discrepancy}&\makecell[c]{Evaluation of distance \\between generated and \\real data using video \\feature extractor}&The same as FID\\
        
        SSIM\cite{FVD}&$SSIM = \frac{(2\mu_x\mu_y + c_1)(2\sigma_{xy} + c_2)}{(\mu_x^2 + \mu_y^2 + c_1)(\sigma_x^2 + \sigma_y^2 + c_2)}$& \makecell{$SSIM$ measures luminance $\mu$, \\contrast $\sigma$, and structure similarity\\ between generated and \\real samples}&\makecell[c]{Evaluation of structural \\similarity to represent human \\perception.}&\makecell{Complex and large \\computation cost}\\
        
        CLIPSIM\cite{clip}&$SIM = \frac{\upsilon_{text} \cdot \upsilon_{video}}{||\upsilon_{text}|| \cdot ||\upsilon_{video}||}$&\makecell{$CLIPSIM$ measures similarity \\$cos(\cdot)$ between text and video \\features extracted by CLIP} &\makecell[c]{Evaluation of similarity \\between text and generated \\data using clip\cite{clip}}&\makecell[c]{Simple calculation, but \\it cannot completely represent \\human perception}\\
        
        PSNR\cite{PSNR}&\makecell{$MSE=\frac{1}{mn}\sum\limits_{i=0}^{m-1}\sum\limits_{j=0}^{n-1}||I(i, j)-K(i, j)||^2$ \\ $PSNR=10\cdot log_10(\frac{MAX_I^2}{MSE})$}&\makecell{$PSNR$ measures the ratio \\of the peak signal energy\\ to the average noise \\energy $MSE$}&\makecell[c]{It compares the \\differences in pixel values \\between two images.}&\makecell[c]{Simple calculation, but it cannot \\ completely represent human\\ perception}\\
        
        \hline
        \multicolumn{4}{l}{Medical Synthetic Video Metrics}\\
        \hline
        
        BmU\cite{sun2024bora}&$BmU = cos(BERT(T_{new}^i), BERT(T_{org}^i))$&\makecell{$BmU$ calculates BERT \\similarity between original \\text prompt $T_{org}$ and generated \\text $T_{new}$ of synthetic videos }&\makecell[c]{Evaluating the degree of \\adherence to prompts within the \\latent space. }&\makecell[c]{Effectiveness has not been \\validated on other medical \\video generation methods.}\\
    \end{tabular}
  \end{center}
\end{sidewaystable}
Evaluation metrics are essential for assessing generation performance. When evaluating natural video generation capabilities, it is important to consider key factors such as generation quality, temporal continuity, and consistency between the condition and the video, e.g., in text-to-video scenarios. To evaluate these aspects, IS~\cite{IS}, FID~\cite{fid}, FVD~\cite{FVD}, SSIM~\cite{ssim}, PSNR, and CLIPSIM~\cite{clip} are widely used for gauging the quality of generated videos.


The current mainstream benchmarks for video generation models include two categories: Text-to-Video (T2V) benchmarks such as VBench~\cite{ji2024t2vbench}, T2VBench~\cite{ji2024t2vbench}, and T2V-Compbench~\cite{t2v-compbench}, and Image-to-Video (I2V) benchmarks such as AIGCBench~\cite{fan2023aigcbench}.

VBench~\cite{lottarini2018vbench} provides 16 evaluation metrics including object identity, motion smoothness, and space relationship. Each dimension is designed as a set of around 100
prompts. VBench~\cite{lottarini2018vbench} has assessed open-source video generation models including ModelScope~\cite{wang2023modelscope}, CogVideo~\cite{hong2022cogvideo}, VideoCrafter-1~\cite{chen2023videocrafter1},and Show-1~\cite{zhang2023show} and closed-source models such as Gen-2~\cite{gen-2} and Pika~\cite{pika}. Among the open-source models, VideoCrafter-1~\cite{chen2023videocrafter1} and Show-1~\cite{zhang2023show}  exhibited notable superiority. While closed-source models excelled in video quality, including aesthetic and imaging quality, certain open-source models surpassed their closed-source counterparts in terms of semantic consistency with user input prompts.

I2V benchmarks, unlike T2V benchmarks, have reference videos in the dataset, 
AIGCBench~\cite{fan2023aigcbench} contains $3928$ samples including real-world video-text and image-text pairs and generated image-text pairs. AIGCBench~\cite{fan2023aigcbench} evaluated both open-source models including VideoCrafter~\cite{chen2023videocrafter1}, and SVD~\cite{stable-video-diffusion} and closed-source models including Pika~\cite{pika} and Gen-2~\cite{gen-2} on $11$ metrics of $4$ dimensions including control-video alignment, motion effect, temporal consistency, and video quality. SVD~\cite{stable-video-diffusion} achieves the best on this benchmark among the open-source models and exhibits results that are comparable to those of closed-source models.

\section{Generative Video Research in Biomedicine}\label{Medical Videos}

Video generation holds significant importance in the medical field, as it can enhance the quality of medical education and improve clinical decision-making. However, it is essential to recognize the substantial differences between medical video data and natural video data.

Medical imaging is typically multimodal. For example, Pathological images typically include multiple staining methods to make various parts of the specimen clearly visible under the microscope, encompassing the structure, morphology, and abnormal changes of cells and tissues. Moreover, distinct staining techniques can differentiate between various tissue components. Medical imaging requires higher contrast than natural images to highlight diagnostic information effectively. Additionally, some medical imaging modalities, such as ultrasound, do not provide depth information and these videos depend on modalities rather than depth cues for control, in contrast to natural videos.

In addition, the generated medical videos usually require a high signal-to-noise ratio to have high diagnostic value~\cite{zhao2024largeDSA}, unlike natural videos, which often focus more on continuity and realism, and do not emphasize the signal-to-noise ratio as much as medical videos do. From existing medical examinations and datasets, such as surgical video recordings~\cite{bawa2021saras, bawa2020esad} and ultrasound imaging~\cite{ouyang2020video}, it can be observed that the subject of imaging is usually located in the center of the video and does not have significant motion, such as the heart only beating rhythmically within a small range. In contrast, natural videos often have significant changes in perspective and object displacement. Therefore, in terms of understanding physical laws, the generation of medical videos is comparatively simpler.

\subsection{Medical Video Dataset and Generation Techniques}
Video generation methods have demonstrated significant promise in the medical field, with the curation of diverse video datasets. These datasets encompass a range of categories, including 1) surgical video datasets; 2) medical imaging video datasets; 3) microscopic video datasets; 4) medical observatory video datasets; 5) medical animation datasets; and 6) medical websites and libraries (see Table~\ref{tab:datasets}). Leveraging these datasets, a variety of methods have been developed, including endoscopic surgery video generation and microscopic video generation. 

\begin{sidewaystable}
\scriptsize
  \begin{threeparttable}[b]
    \centering
    \caption{Medical video datasets, including minimally invasive surgery (MIS) videos, interventional surgery (IS) videos, real-time MRI videos, videos under a microscope, medical observatory videos, and medical animations.}
    \begin{tabular}{c|c|c|c|c|c|c|c|c} 
      \textbf{Dataset} & \textbf{Type}&\textbf{Category} & \textbf{Tasks}   &\textbf{Video} & \textbf{Label} &\textbf{Class}& \textbf{Avg.}&\textbf{Resolution}\\
      \hline
MESAD-Real\cite{bawa2021saras, bawa2020esad} & MIS &prostate& Action Cls., Det.&F:23366&Class, Bbox&21&-&$720\times756$\\
MESAD-Phantom\cite{bawa2021saras, bawa2020esad}& MIS &prostate &Action Cls., Det.&F:22609&Class, Bbox&14&-&$720\times756$\\
SurgicalActions160\cite{DBLP:journals/mta/SchoeffmannHKPM18}& MIS &gynecology &Action Cls. &160&Class&16&-&-\\
Cataract-21\cite{DBLP:conf/mmm/PrimusPTMEBS18}&MIS&Ophtha.&Action Cls.&21&Class&10&-&-\\
Cataract-101\cite{DBLP:conf/mmm/PrimusPTMEBS18}&MIS&Ophtha.&Action Cls.&101&Class&10&-&-\\
IrisPupilSeg\cite{DBLP:conf/isbi/SokolovaTSPS20}&MIS&Ophtha.&Iris pupil Seg.,&35&Mask&-&-&$540\times720$\\
CatInstSeg-Manu\cite{DBLP:conf/cbms/FoxTS20}&MIS&Ophtha&Inst. Seg.&F:843&Class, Mask, bbox&11&-&-\\
CatInstSeg-Auto\cite{DBLP:conf/cbms/FoxTS20}&MIS&Ophtha.&Inst. Seg.&F:4738&Class, Mask, bbox&15&-&-\\
CatRelComp-1\cite{DBLP:conf/mm/GhamsarianATTS20}&MIS&Ophtha&Idle Rec.&F:22000&Class&22&-&-\\
CatRelComp-2\cite{DBLP:conf/mm/GhamsarianATTS20}&MIS&Ophtha&Cornea, Inst. Seg.&F:478&Mask&11&-&-\\
CatRelDet\cite{DBLP:conf/icpr/GhamsarianTPSS20}&MIS&Ophtha&Surgery Phase Cls.&2200&Class&4&3s&-\\
LensID-1\cite{DBLP:conf/miccai/GhamsarianTPSES21}&MIS&Ophtha&Surgery Phase Cls.&100&Class&2&3s&-\\
LensID-2\cite{DBLP:conf/miccai/GhamsarianTPSES21}&MIS&Ophtha&Lens Seg.&27&Mask&-&-&-\\
PitVis\cite{PitVis}&MIS&Pituitary &Seg. Cls.& 33 & Mask&Act:14,Tool:18&72.8min&720*720\\
SurgToolLoc\cite{zia2023surgical}&MIS&laparoscope&Tool Local. Cls.&24695&Class, bbox&14&30s&$1280\times720$\\
CholecT50\cite{nwoye2022rendezvous}&MIS&laparoscope&Triplet det. Rec.&50&Class, bbox&-&-&-\\
CholecT80\cite{twinanda2016endonet}&MIS&laparoscope&Det. Rec.&80&class,bbox&Actions:7,Tools:7&-&-\\
CholecT40\cite{nwoye2020recognition}&MIS&laparoscope&Triplet Det. Rec.&40&Class,Bbos&6&-&-\\
SAR-RARP50\cite{psychogyios2023sar}&MIS&laparoscope&\makecell{Action Rec.; \\Inst. Seg. Cls.}&50&Class,Mask&Act:8, Inst:9&-&$1920\times1080$\\
FetReg\cite{bano2021fetreg, bano2023placental}&MIS&-&Placental Seg., Register&18&Mask&8s&-&-\\
PETRAW\cite{huaulme2022peg}&MIS&-&Workflow Rec.&-&Class, Mask&\makecell{Phases:2;Stages:12;\\Actions:6}&-&$1920\times1080$\\
MISAW\cite{HUAULME2021106452}&MIS&-&Workflow Rec.&-&Class&\makecell{Phases:2;Stages:2,;\\Actions:17}&-&$960\times540$\\
ROBUST-MIS\cite{RO2021101920}&MIS&rectal&Surgery Cls.;Inst. Seg&-&Class, Mask&3&-&-\\
HeiChole\cite{wagner2023comparative}&MIS&-&\makecell{Workflow Rec.;\\Action Cls.; Tool Cls.}&30&Class&\makecell{Phases:7;\\Actions:4,Tools:20}&-&-\\
SWAS\cite{SWAS}&MIS&-&\makecell{Workflow Rec.;\\ Inst. Cls.}&42&Class&Phases:14,Tools:12&-&-\\
RSS\cite{allan20202018}&MIS&Kidney&Inst. Seg.&16&Mask&8&-&-\\
CATARACT\cite{ALHAJJ201924}&MIS&Ophtha.&Tool Cls.&100&Class&21&-&$1920\times1080$\\
RIS\cite{allan20192017}&MIS&-&Inst. Seg.&18&Mask&-&-&-\\
KBD\cite{hattab2020kidney}&MIS&Kidney&Kidney Det.&15&Mask&-&-&-\\
ICT\cite{bodenstedt2018comparative}&MIS&-&Inst. Seg.&34&Mask,Bbox&-&-\\
AOD\cite{7840040}&MIS&Intestine&Polyp Seg.&49&Mask&-&-&-\\
Endoscapes-CVS201\cite{murali2023endoscapes}&MIS&Laparoscope&Critical View of Safety&201&Grade&3&-&-\\
Endoscapes-bbox201\cite{murali2023endoscapes}&MIS&-&Tissue Tool Det.&201&Bbox&6&-&-\\
Endoscapes-Seg201\cite{murali2023endoscapes}&MIS&-&Tissue Tool Seg.&201&Mask&-&-&-\\
SSG-VQA\cite{SSG-VQA}&MIS&Laparoscope&VQA&25k&Text&-&-&-\\
MultiBypass140\cite{lavanchy2024challenges}&MIS&Laparoscope&Workflow Rec.&140&Class&P:12,St:45&-&-\\
Colonoscopic\cite{Colonoscopic}&MIS&Colon&Cls.&76&Class&3&-&$340\times256$\\
Kvasir-Capsule\cite{Kvasir-Capsule}&MIS&Digestive Tract&Cls.&117&Class&14&-&$336\times336$\\
DV-MuRC\cite{zhao2024largeDSA}&IS&-&Generative&F:3m&Mask&-&-&$489\times489$,$512\times512$, ...\\
CAMUS\cite{leclerc2019deep}&Ultrasound&Heart&Seg.&F:1000&Mask&4&-&-\\
Echonet\cite{ouyang2020video}&Ultrasound &Heart&-&10036&Clinical Info&-&-&$112\times112$\\
BUV dataset\cite{lin2022new}&Ultrasound &Breast&Lesion Det.&-&Class, Bbox&2&-&-\\
Uliver\cite{de2013Uliver}&Ultrasound &Liver&Track&7&-&-&-&$500\times480$\\
2dRT\cite{lim2021multispeaker-RTMRI}&RT-MRI&-&-&-&Audio&-&-&$84\times84$\\
Cell Track\cite{mavska2023cell}&Micro&Hela Cell&Track&2&Mask&-&-&$700\times1100$\\
Tryp\cite{anzaku2023tryp}&Micro&Parasites&Det.&114&Bbox&-&-&$1360\times1024$\\
Yeast\cite{yeast-dataset}&Micro&Yeast&Cls.&2417&Class&14&-\\
Embryo\cite{embryo-dataset}&Micro&Embryo&Cls.&704&Class&16&-&$500\times500$\\
CTMC-v1\cite{anjum2020ctmc}&Micro&Cell&Track&86&bbox&-&-&$320\times400$\\
VISEM\cite{haugen2019visem}&Micro&Spermatozoa&Evaluation&85&Text&-&-&$640\times480$\\
PURE\cite{stricker2014non}&Medical Observatory Videos&Face Video&Pulse Rate Est.&60&Pulse Rate&-&60s&$640\times480$\\
IMVIA-NIR\cite{BENEZETH2024106598rppg}&Medical Observatory Videos&Face Video&Pulse Rate Est.&20 &Pulse Rate &- &- &$1280\times1024$ \\
MERL-RICE\cite{nowara2020near}&Medical Observatory Videos&Driving&Pulse rate Est.&18 &Pulse Rate &- &- &$640\times640$ \\
TokyoTech-NIR\cite{maki2019inter}&Medical Observatory Videos&Face Video&Pulse Rate Est.&9 &Pulse Rate &- &180s &$640\times480$ \\
Video-EEG\cite{lin2024vepinet}&Medical Observatory Videos&Movements&Epilepsy Diag.&191025 &Class &2 &4s &$1920\times1080$ \\
SimSurgSkill\cite{zia2022objective}&3D Animation&Surgery&Inst. Cls. Det.&-&Class, Bbox&-&-&$1280\times720$\\
SurgVisDom\cite{zia2021surgical}&3D Animation&Surgery&Surgical Task Cls.&59&Class&3&-&$1280\times720$\\
    \end{tabular}
    \label{tab:datasets}
    \begin{tablenotes}
     \item[1] Abbreviations: Ophtha.-Ophthalmology; Cls.-Classification; Seg.-Segmentation; Local.-Localization; Det.-Detection; Est.-Estimation; Inst.-Instrument; F.-Frames
   \end{tablenotes}
  \end{threeparttable}
\end{sidewaystable}

\begin{table}[h!]
\scriptsize
  \begin{center}
    
    \caption{Video generation methods in medical domain.}
    \begin{tabular}{c|c|c|c|c} 
      \textbf{Domain} & \textbf{Category} &\textbf{Model}&\textbf{Architecture}&\textbf{Conditions}\\
      \hline
      \multirow{3}{*}{Surgery Video Generation}&IS&GenDSA\cite{zhao2024largeDSA}&Optical Flow&Low FPS Video\\ 
      &MIS&Endora\cite{li2024endora}&Diffusion Model&Unconditional\\
      &MIS&Surgen\cite{cho2024surgen}&Diffusion Model&Surgical Description\\
      \hline
      \multirow{7}{*}{Medical Imaging Video Generation}&Ultrasound&Nguyen Van Phi\cite{van2023echocardiography}&Diffusion Model&Cardiac Structure Mask\\
      &Ultrasound&Hardrein\cite{reynaud2023feature}&Diffusion Model&\makecell{Echocardiogram Image\\Clinical Parameters}\\
      &Ultrasound&Echonet-Syn\cite{reynaud2024echonet}&Diffusion Model&\makecell{Privacy-preserving Heart\\Clinical Parameters}\\
      
      &Ultrasound&Pellicer AO\cite{pellicer2024generation}&Diffusion Model&Unconditional\\
      &Ultrasound&Jiamin Liang\cite{liang2022weakly}&GAN&Motion\& Key Point\\
      &Ultrasound&ECM\cite{yu2024explainable}&Diffusion Model&Cardiac Motion\\
      &Ultrasound&HeartBeat\cite{zhou2024heartbeat}&Diffusion Model&\makecell{Sketch Mask\\Skeleton Optical flow\\Echocardiogram Image}\\
      &FFA&Fundus2Video\cite{fundus2video}&GAN&\makecell{Fundus Image\\Knowledge Mask}\\
      \hline
      \multirow{3}{*}{Microscopic Video Generation}&Cell Track&BVDM\cite{yilmaz2024annotated}&Diffusion Model&Cell Mask\\
      &Yeast\&Embryo&P\'erez PC\cite{Generation-evaluation}&\makecell{Diffusion Model\\GAN}&Unconditional\\
      &Embryo&EmbryoTgan\cite{celard2023study}&GAN&Unconditional\\
      \hline
      Medical Observatory Video Generation & Near-infrared Videos&Yannick\cite{9150894}&\makecell{dichromatic \\reflection model}&rPPG Signal\\
      \hline
      Medical Animation& 3D Animation&SyntheticColon\cite{jagtap2022automatic}&3D Model&Unconditional\\
      \hline
      Multimodal Video Generation&\makecell{MRI MIS\\Cell Ultrasound}&Bora\cite{sun2024bora}&Open Sora\cite{opensora}&Medical Description
      
    \label{tab:medical video dataset}
    \end{tabular}
  \end{center}
\end{table}

\subsubsection{Surgical Video Generation}\label{surgery video}

Surgical video datasets are often drawn from different surgical scenarios, including open surgery, minimally invasive surgery, small-incision surgery, endoscopic surgery, robotic surgery, and interventional operation.  For simple classification, the video datasets from these $6$ kinds of surgeries are categorized into $3$ kinds of video datasets: open surgery (OS) video dataset, interventional surgery (IS) video dataset, and minimally invasive surgery (MIS) video dataset.

\textbf{OS Video Generation} OS is a traditional type of surgery that exposes the surgical site through a large incision, allowing the doctor to operate directly under direct vision. Therefore, it is more suitable for large-scale operations with complex and difficult operations. Few open surgical videos are curated into structured datasets, most of which are included in online websites (see Table \ref{Online Resources}). Nonetheless, there is currently no video generation technology developed for open surgeries.

\textbf{IS Video Generation} IS, guided by imaging equipment, introduced instruments such as guide wire and catheters into the human body through minimal incisions and the body's natural orifices to diagnose and treat diseases. 

GenDSA~\cite{zhao2024largeDSA} proposes a flow-mask-based model MoStNet to interpolate between frames based on DV-MuRC~\cite{zhao2024largeDSA} containing 3 million digital subtraction angiography frames from $27117$ patients. It extract motion-structure information by fusing multi-scale features of two frames and predict flow between two of them. This method is capable of generating a whole digital subtraction angiography video using only $\frac{1}{3}$ of total frames, helping to reduce the radiation exposure for patients while lowering the cost of detection.

\textbf{MIS Video Generation} MIS is a surgical approach conducted through smaller incisions or the body's natural orifices, aiming to achieve the best therapeutic outcomes with minimal trauma. It mainly uses modern medical instruments such as laparoscope, thoracoscope, and arthroscope to enter the human body through tiny incisions or natural cavities for fine surgical operations. 

Endora~\cite{li2024endora}, trained on Colonoscopic~\cite{Colonoscopic}, Kvasir-Capsule~\cite{Kvasir-Capsule}, and CholecT~\cite{nwoye2022rendezvous}, generates medical videos that simulate minimally invasive surgery. It introduces a diffusion backbone of transformer with spatial and temporal blocks to deal with endoscopic videos and extracts a prior from real video by DINO encoder~\cite{zhang2022dino} as a guide to control video synthesis using a conditional mechanism of Pearson correlation. It shows decent performance with regarded to video generation metrics such as FVD, FID, and IS. Surgen~\cite{cho2024surgen}, trained on Cholec80~\cite{twinanda2016endonet}, designs its diffusion model based on DiT~\cite{diffusion-transformer} conditioned on text to generate high-quality surgical videos.

Apart from the surgical datasets used in Endora and Surgen~\cite{Colonoscopic, Kvasir-Capsule, twinanda2016endonet, nwoye2022rendezvous}. MIS also provides a large scale of datasets for surgery video generation model training, including laparoscope surgery~\cite{SurgToolLoc-instrument-local, nwoye2020recognition, psychogyios2023sar, murali2023endoscapes, SSG-VQA, lavanchy2024challenges}, ophthalmological surgery~\cite{DBLP:conf/mmm/PrimusPTMEBS18, DBLP:conf/isbi/SokolovaTSPS20, DBLP:conf/cbms/FoxTS20,  DBLP:conf/mm/GhamsarianATTS20, DBLP:conf/icpr/GhamsarianTPSS20, DBLP:conf/miccai/GhamsarianTPSES21}, and others~\cite{bano2021fetreg, bano2023placental, huaulme2022peg, HUAULME2021106452, RO2021101920, wagner2023comparative, SWAS, allan20202018, ALHAJJ201924, allan20192017, hattab2020kidney, bodenstedt2018comparative, 7840040}.

\subsubsection{Medical Imaging Video Generation}\label{Medical Imaging Video}

Medical imaging refers to the use of contrast techniques to obtain images of the internal structures and functions of the human body, such as CT, MRI, X-ray, and ultrasound. Given the scarcity of medical imaging data and the radiation associated with certain medical imaging examinations, generating corresponding videos holds applied value and clinical benefits. Some efforts~\cite{van2023echocardiography, reynaud2023feature, reynaud2024echonet, pellicer2024generation, liang2022weakly, yu2024explainable, zhou2024heartbeat, fundus2video} have been made in this direction.

Nguyen Van Phi et al. ~\cite{van2023echocardiography} proposed a conditional diffusion model for echocardiography video synthesis guided by semantic mask. It uses spatial adaptive normalization~\cite{wang2022semantic} to introduce semantic conditions into the denoising process. Trained on CAMUS~\cite{leclerc2019deep} dataset, this model is able to produce realistic echocardiography videos consistent with semantic segmentation map.

Hardrien et al.~\cite{reynaud2023feature} generate ultrasound video from a single image and left ventricular ejection fraction (LVEF) score based on EchoNet~\cite{ouyang2020video}. To better synthesize video data from a single image and interpretable clinical data, Eclucidated Diffusion Model(EDM)~\cite{karras2022elucidating} is applied. Additionally, further experiments proved that proposed method has fine-grained control of specific properties like LVEF leading to precise data generation. Based on EchoNet~\cite{ouyang2020video}, Hadrien et al. also developed a framework based on LVDM to generate longer echocardiogram videos at near real-time speeds~\cite{reynaud2024echonet}. Besides, EchoNet-Synthetic dataset was proposed, with competitive performance and quality as real data. Another method of generating high-quality echocardiogram videos proposed by Alexandre et al.~\cite {pellicer2024generation} is capable of generating echocardiograms of four different views. 

In the field of ophthalmology, Fundus2Video~\cite{fundus2video} attempts to generate Fundus Fluorescein Angiography (FFA) from Color Fundus (CF). Based on an in-house CF-FFA dataset, It auto-regressively trains a GAN to generate FFA sequence from CF guided by a knowledge mask. In order to align the three of them with each other, including input CF, knowledge mask, and FFA sequence, it designed knowledge-boosted attention and knowledge-aware discriminators for specific supervision on lesion regions. Experiments show that it successfully addresses challenges in lesion generation and pixel misalignment.

Existing open-source medical imaging video datasets are mainly from two categories: 1) ultrasound datasets~\cite{reynaud2024echonet, lin2022new, de2013Uliver, leclerc2019deep, liang2022weakly, yu2024explainable} and 2) real-time MRI datasets~\cite{lim2021multispeaker-RTMRI}. While these datasets form the basis for generating related imaging videos, publicly available data is still relatively scarce, especially for other imaging modalities such as CT  and FFA.

\subsubsection{Microscopic Video Generation}
Microscopic videos record the biomedical behaviors under the microscope, primarily the activity of microorganisms, including 1) microorganisms' morphology, i.e., morphological changes of microorganisms at different time points, such as division, migration, and morphological changes; 2) microorganisms' behavior, i.e., behavior of microorganisms in a specific environment, such as microorganisms interactions, and responses to stimuli; and 3) microorganisms' labeling, i.e., dynamic changes of specific molecules or structures within cells through techniques such as fluorescent labeling. Cell Track~\cite{mavska2023cell}, Tryp~\cite{anzaku2023tryp}, Yeast~\cite{yeast-dataset}, Embryo~\cite{embryo-dataset}, CTMC-v~\cite{anjum2020ctmc}, and VISEM~\cite{haugen2019visem} record various kinds of behaviors of various types of microorganisms such as Hela cell and yeast.

Trained on Hela Cell Track~\cite{mavska2023cell}, BVDM~\cite{yilmaz2024annotated} consists of two parts: DDPM~\cite{NEURIPS2020_4c5bcfec} and VoxelMorph~\cite{balakrishnan2019voxelmorph} for flow field prediction. Living cell video frames are trained on DDPM~\cite{NEURIPS2020_4c5bcfec} for image generation, and two consecutive masks are trained on flow prediction model~\cite{balakrishnan2019voxelmorph} for finding flow field between two consecutive images. During inference stage, the diffusion model generates texture based on the first mask of cells. For the rest of frames, flow prediction from VoxelMorph~\cite{balakrishnan2019voxelmorph} is applied to the output from previous iteration, and the result is fed to DDPM to generate the next frame. This framework tackles the scarcity of annotated real living cell datasets and, training on the synthetic data generated by BVDM~\cite{yilmaz2024annotated} has proved superior performance compared to training with a limited amount of real data.

Pedro Celard Pérez et al.~\cite{Generation-evaluation} compared video generation performance based on yeast image sequence~\cite{yeast-dataset} and embryo video dataset~\cite{embryo-dataset} with video diffusion model (VDM)~\cite{he2022latent} and Temporal GANv2 (TGANV2)~\cite{saito2017temporal}. Results showed that with regard to $64\times64$ and $128\times128$ image and video generation, TGANv2~\cite{saito2017temporal} have better performance than VDM~\cite{he2022latent} in FID~\cite{fid} and FVD~\cite{FVD}.

\subsubsection{Medical Observatory Video Generation}
Medical observatory videos consistently document biological behaviors, including morbidity and behavioral records that reflect certain medical situations, such as the stages of epilepsy onset~\cite{lin2024vepinet} and cardiac dynamics~\cite{stricker2014non, BENEZETH2024106598rppg, nowara2020near, maki2019inter}. Due to the clinical value of such data and its paired vital signals, such as remote photoplethysmography (rPPG) and electroencephalogram(EEG), and the scarcity of paired video-signal data, the synthesis of such data is worth exploring.

Because capturing videos using infrared cameras to reflect rPPG signals is costly, employing synthetic methods to expand such datasets is hence a cost-effective approach. Yannick et al.~\cite{BENEZETH2024106598rppg} proposed a method to generate videos using synthetic rPPG signals~\cite{9150894, 8857081}. They first generate the rPPG signals, then generate the spatial, channel, and motion dimensions of the video based on the rPPG signals, and finally integrate them into a complete video. 

In addition to the datasets~\cite{stricker2014non, BENEZETH2024106598rppg, nowara2020near, maki2019inter} used in the algorithm proposed by Yannick et al~\cite{BENEZETH2024106598rppg}, EEG signals can also guide video generation. Video-EEG proposed by VepiNet~\cite{lin2024vepinet} contains $191025$ video-EEG segments from $484$ patients, providing a foundation for generating epilepsy videos from EEG signals. 

\subsubsection{Medical Animation and Generation}
Medical animation offers a compelling way for the public to learn about medicine, featuring animations that explain complex medical concepts and 3D simulators that demonstrate, e.g., surgical procedures

SyntheticColon~\cite{jagtap2022automatic}was not trained on existing 3D animation video datasets. It renders a 3D model and generates videos by making a camera pass through the intestinal model.

Both SimSurgSkill~\cite{zia2022objective} and SurgVisDom~\cite{zia2021surgical} contain VR videos for training and surgical videos for testing to tackle domain adaption challenges. Taking SimSurgSkill as an example, it includes 157 VR videos at 30 frames per second with a resolution of $1920\times1080$. Generating 3D animated surgical videos can contribute to the promotion of surgical knowledge. With the advances in video generation, it may become a fast and effective approach that is comparable with current VR and 3D rendering approaches in medical animation. 

\subsubsection{Multimodal Video Generation}

The generation of multi-modal medical videos implies that the model can simultaneously generate medical data from different modalities, such as surgical videos, ultrasound videos, and microscopic videos. This type of generative model is commonly referred to as an all-in-one generation model.

Bora~\cite{sun2024bora} is the first diffusion model designed for text-guided multimodal biomedical video generation, fine-tuned on a new large-scale medical video corpus, including paired text-video data of endoscope~\cite{Colonoscopic, Kvasir-Capsule, nwoye2022rendezvous}, cardiac ultrasound~\cite{de2013Uliver, ouyang2020video}, real-time MRI~\cite{lim2021multispeaker-RTMRI} and cellular visualization~\cite{anzaku2023tryp, anjum2020ctmc, haugen2019visem}. The video captions are generated by LLM~\cite{achiam2023gpt} with background information such as technical documents and research papers. Fine-tuned on Sora~\cite{sora}, Bora can~\cite{sun2024bora} not only understand the caption details but generate realistic videos, outperforming other video generation techniques~\cite{pika, pixelverse, gen-2, wang2023lavie, wang2023modelscope}.

Despite that, it is worth considering that, as mentioned in AMIR~\cite{allinone}, due to the differences between medical and natural images, the implementation of an all-in-one model for medical applications faces greater difficulties and challenges, even though the medical all-in-one model increases medical analyzing efficiency. For instance, the restoration of natural images involves reversing various image degradations to the original RGB image distribution, whereas the restoration of medical images not only requires the repair of image degradation but also the restoration of the image to different original medical distributions, which may lead to interference between different tasks, significantly increasing the difficulty of restoration. This same situation applies to Bora which generates multiple modalities of videos simultaneously, which may greatly increase the complexity. Since there is no clear evidence that an all-in-one model~\cite{sun2024bora} is a better design strategy compared with domain-specific models such as Endora, it is difficult to verify whether an all-in-one model in medical video generation will affect the quality of generation. 

\subsubsection{Medical Video Libraries and Websites}\label{Libraries}
This section summarizes medical video sources from online platforms such as YouTube and various video libraries (see Table \ref{Online Resources}), including surgery videos especially open surgeries and videos explaining medical knowledge, complementing the curated medical video datasets found in the literature. 

For example, Stryker~\cite{Stryker-Surgical-Videos} contains $23$ complete surgical procedures, with each video lasting several tens of minutes from $5$ kinds of surgical categories including facial trauma, orthognathic surgery, facial reconstruction, neurosurgery, and temporomandibular joint of $19$ surgical techniques. Vit-bult~\cite{Eyes-Surgical-video-dataset} provides ophthalmic surgeries of $5$ kinds, including intraocular lens surgery, macular surgery, pediatric surgery, trauma surgery, and uveity surgery videos.

\begin{table}[h!]
\footnotesize 
  \begin{center}
    \caption{Online medical video libraries and websites}
    \begin{tabular}{c|c|c|c} 
      \textbf{Library} &\textbf{Type}& \textbf{Library} &\textbf{Type} \\
      \hline
Omnimedicalsearch~\cite{omnimedicalsearch}&Surgery, Medical Knowledge&SGS Library\cite{SGS}&Surgery, Medical Knowledge\\
Spinalsurgicalvideo\cite{Spine-surgery-library}&Surgery&Coronary Bypass Surgery\cite{Coronary-Bypass-Surgery}&Surgery\\
CarpalTunnel\cite{Carpal-Tunnel}&Surgery&TotalKneeReplacement\cite{Total-Knee-Replacement-Surgery}&Suregry\\
TotalHipReplacement\cite{Total-Hip-Replacement-Surgery}&Surgery&Tracheostomy\cite{Tracheostomy}&Surgery\\
Thyroidectomy\cite{Thyroidectomy}&Surgery&Endoscopic Sinus Surgery\cite{Functional-Endoscopic-Sinus-Surgery}&Surgery\\
 Stryker Surgical Videos\cite{Stryker-Surgical-Videos}&Surgery&MED EL Video library\cite{MED-EL-Surgical-Video-library}&Surgery\\
 ACS Online Library\cite{ACS-Online}&Surgery&HIF library\cite{House-Institute-Foundation-surgical-video-library}&Surgery\\
 PlasmaJet Video Library\cite{PlasmaJet-video-library}&Surgery&Eyes Surgical video\cite{Eyes-Surgical-video-dataset}&Surgery\\
Surgical video library\cite{Surgical-video-library-for-education}&Surgery, Medical Knowledge&YouTube Medcram\cite{Medcram}&Medical Knowledge\\
YouTube Osmosis\cite{osmosis}&Medical Knowledge& lecturiomedical\cite{lecturiomedical}&Medical Knowledge\\
 DoctorNajeeb\cite{DoctorNajeeb}&Medical Knowledge& NinjaNerdOfficial\cite{NinjaNerdOfficial}&Medical Knowledge\\ armandohasudungan\cite{armandohasudungan}&Medical Knowledge&TheMDJourney\cite{TheMDJourney}&Medical Knowledge\\ZeroToFinals\cite{ZeroToFinals}&Medical Knowledge&StrongMed\cite{StrongMed}&Medical Knowledge\\
geekymedics\cite{geekymedics}&Surgery, Medical Knowledge& nucleusmedicalmedia\cite{nucleusmedicalmedia}&Medical Knowledge\\Anatomyzone\cite{Anatomyzone}&Medical Knowledge& SpeedPharmacology\cite{SpeedPharmacology}&Medical Knowledge\\
Atrial-Fibrillation\cite{Atrial-Fibrillation}&Medical Knowledge&CAB Graft\cite{Coronary-Artery-Bypass-Grafting}&Medical Knowledge
    \end{tabular}
    \label{Online Resources}
  \end{center}
\end{table}

\subsection{Medical Video Metrics and Benchmarks}
Metrics such as IS~\cite{IS}, FID~\cite{fid}, and FVD~\cite{FVD} have been widely used in evaluating the performance of generating natural videos (see section ~\ref{Natural Video Metrics}). However, these traditional metrics may not fully capture the nuances and complexities inherent in medical data, highlighting the need for tailored biomedical metrics.  Biomedical Understanding (BmU) proposed by Bora~\cite{sun2024bora}  evaluates biomedical video generation performance, which calculates BERT similarity between the origin biomedical prompt and generated text of synthetic videos to evaluate the degree of adherence to prompts within the latent space. However, in this metric, BERT is trained on general domain data, and therefore its suitability for medical analysis would be strengthened if fine-tuned with medical data.

\section{Future Applications}\label{Application}

Although there are currently many video generation technologies, only a few of them have been applied to medical scenarios, leaving majority of scenarios unexplored. Therefore, this chapter will discuss potential application scenarios for medical video generation and provide a comprehensive analysis of the technical prerequisites and existing limitations. Specifically, the chapter will examine six key application areas: 1) medical education; 2) patient-facing applications; 3) public health; 4) integration with AI models; 5) diagnostic assistance; and 6) biomedical simulations.

\subsection{Medical Education}

Video is a more effective medium for imparting knowledge compared to static images and text. Due to the limitations of online medical video resources, customizable synthetic medical videos have a broader range of practical applications, particularly beneficial in fields like surgical and clinical education where tailored content is essential for effective learning~\cite{wei2024medco}.
\begin{figure}[htb]
    \includegraphics[page=1, width=1\columnwidth]{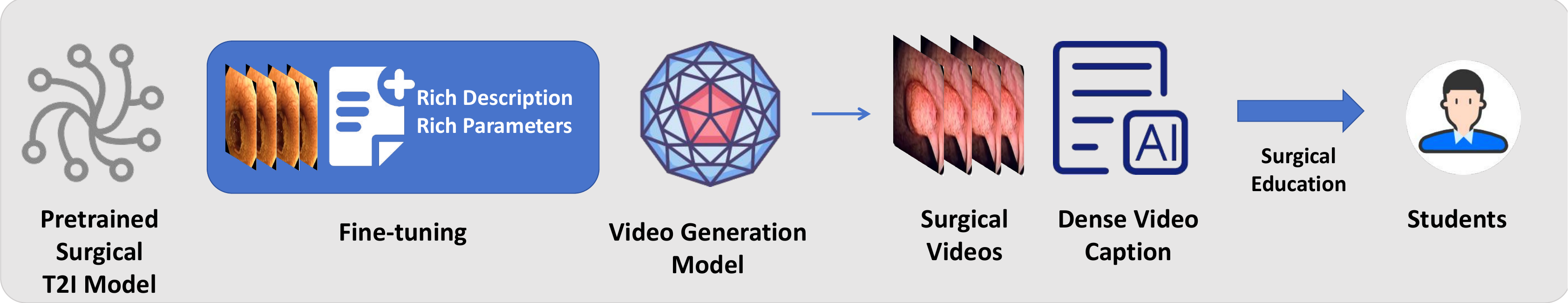}
    \caption{Video generation model for medical education, with surgical education as an example.}
    \label{fig:Architecture}
\end{figure}
\subsubsection{Surgical Education}

Generative models hold the potential to enhance surgical education by producing synthetic videos that explain surgical concepts and simulate surgical procedures~\cite{lam2024foundation} to provide a more diverse range of surgical guidance in a more dynamic and effective manner for students. As a potential simulator of medical surgery, surgical video generation models will be capable of comprehensively understanding and mastering the various stages of complicated surgical procedures, generating surgical scenes, which requires surgical video generation models to establish a robust mapping between this knowledge and corresponding visual representations, and consequently, possess both text-to-video and video-to-text generation capabilities.

Training this surgical video generator requires a large-scale video dataset with rich visual and textual pairings, including textual descriptions of the videos and relevant surgical parameters such as categories, locations, and angles of surgical instruments.

When generating a video, a commentary for the surgical video can also be produced simultaneously for better education. Surgical video captioning techniques including SwinMLP-TranCAP~\cite{xu2022rethinking} and SCA-Net~\cite{chen2023surgical} are able to caption from surgical videos. Dense Video Captioning~\cite{zhou2024streaming} is also a suitable captioning method for surgical video captioning.

\subsubsection{Other Applications}

In addition to its application in surgical education, video generation techniques can be effectively utilized in various medical contexts, such as generating educational materials, providing videos to explain medical concepts and knowledge.

\textbf{Educational Video Materials} There are numerous explanations of medical textbooks on YouTube that facilitate more efficient learning. However, these explanations often fail to cover all aspects of the knowledge. Therefore, video generation models can be utilized to produce explanatory videos targeted at specific knowledge points, aiding in the learning process. Unlike the intricate realism required for surgical education scenarios, applications in visual educational materials do not necessitate highly detailed scenes, thereby lowering the complexity of generation while enhancing practical feasibility. 

\subsection{Patient-facing Application}

Patient-facing applications are digital tools or platforms designed specifically for patients to interact with healthcare services. Virtual video consultation is one of the key applications of video generation in this area.
\begin{figure}[htb]
    \includegraphics[page=1, width=1\columnwidth]{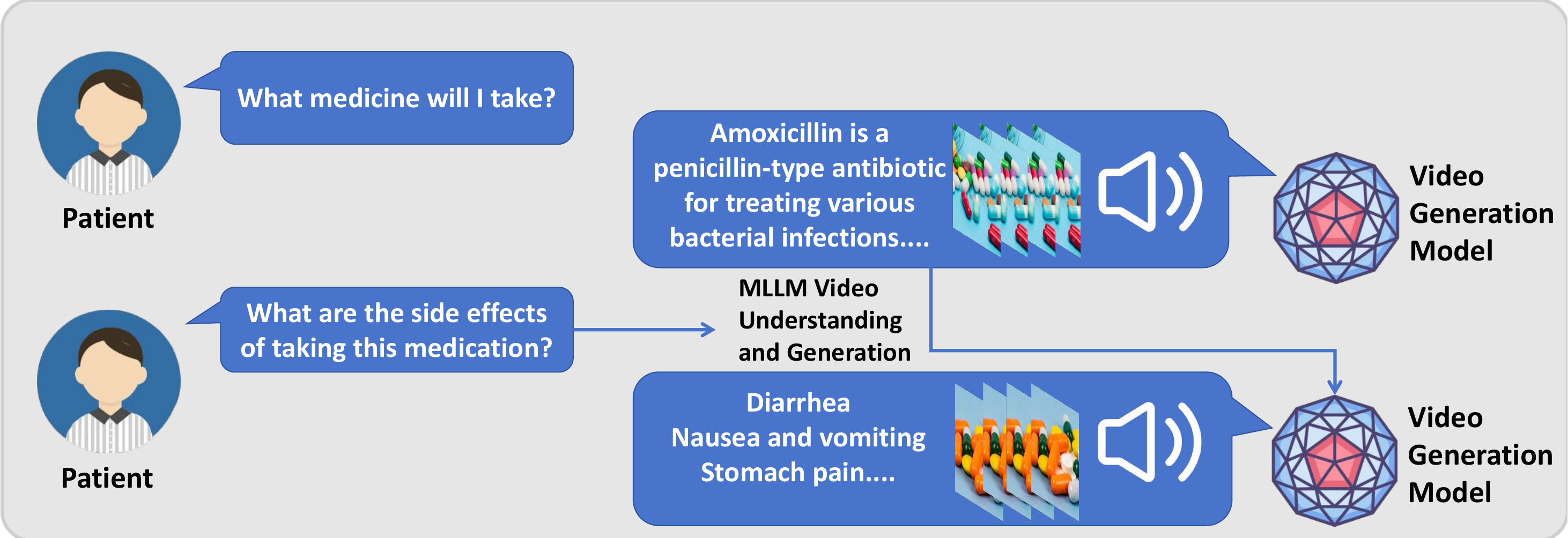}
    \caption{Video generation model for patient-facing application, with video-based virtual consultation as an example.}
    \label{fig:Architecture}
\end{figure}
\subsubsection{Virtual Video Consultation}

Virtual consultations provide an effective solution for urgent medical needs when immediate physician availability is lacking. As a dynamic multimedia medium, video conveys medical information more intuitively and vividly than traditional text or image formats~\cite{tencent-ipc, baidu01}. The consultation system is capable of generating videos to assist patients through dialogue, and by comprehending the semantic context of preceding conversations, they can produce more accurate and appropriate video responses. This offers significant advantages in enhancing communication and understanding between healthcare providers and patients.

For effective dataset preparation, ensuring diversity is paramount to cater to the spectrum of patient inquiries. Similar to the question-answer (QA) dataset used in LLMs such as WiKiQA~\cite{yang2015wikiqa}, the dataset for training the video-based consultation system should adopt a video QA format, specifically the question-video framework. Thoughtfully crafted prompts must be curated to meet the system's requirements.

Similar to LLMs~\cite{touvron2023llama, achiam2023gpt}, the video-based consultation system is supposed to have the ability to multi-turn conversation, which is, video can be generated not only based on previous conversations, but based on the previous videos. For instance, patients should have the ability to critique previous explanations, prompting system corrections. Current works such as VideoChatGPT~\cite{maaz2023video} have the capability of understanding video based on Multimodal LLM. Previous works have attempted to edit images based on Multimodal LLMs~\cite{fu2023guiding, huang2024smartedit}, demonstrating the potential of video editing capability.

The video generation model presents a promising solution for enhancing medical consultations. Particularly, when patients are preparing to embark on specific medication regimens, treatments, or surgical procedures, the utilization of a virtual treatment guide can offer elucidating videos that surpass the efficacy and utility of verbal or written instructions. While existing methodologies can generate videos based on a well-curated dataset utilizing current techniques~\cite{sora, stable-video-diffusion}, the development of a comprehensive system for multi-turn conversation consulting and treatment necessitates further investigation and refinement.

\subsection{Public Health}

Medical video generation holds immense potential across various public health contexts. Its applications extend beyond merely enhancing the efficiency of disseminating health knowledge to bolstering public health awareness such as promoting healthy diet~\cite{qiu2023egocentric, lo2024dietary}.
\begin{figure}[htb]
    \includegraphics[page=1, width=1\columnwidth]{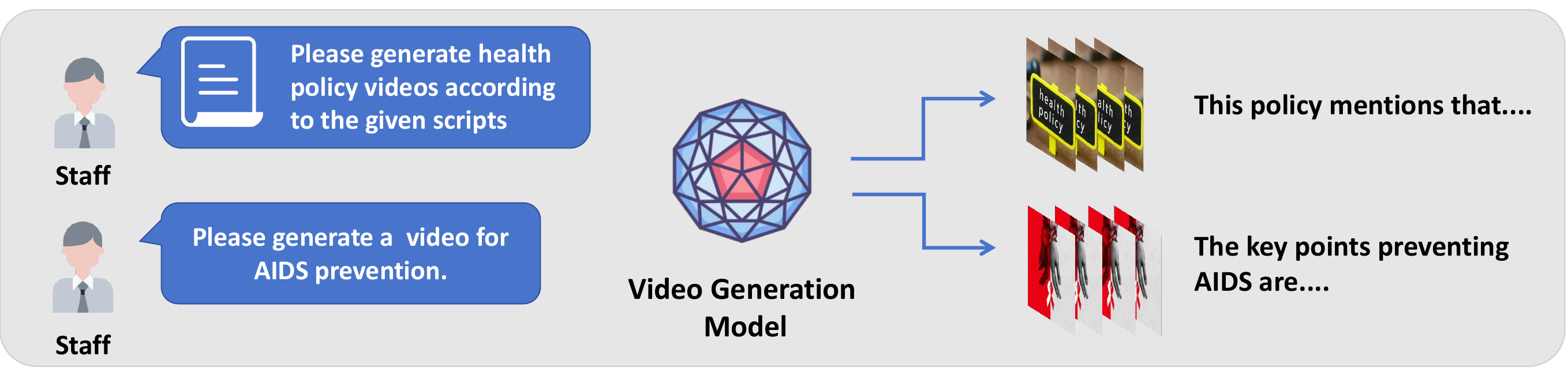}
    \caption{Video generation model for public health, with public health promotion as an example.}
    \label{fig:Architecture}
\end{figure}
\subsubsection{Public Health Promotion} 
By creating medical videos focused on the prevention, symptom identification, and treatment methods for common diseases such as influenza, AIDS, and diabetes, the government can effectively disseminate health knowledge to the public and enhance disease prevention and control efforts. Additionally, producing videos that interpret public health policies can further aid the public in understanding these important initiatives.

Knowledge dissemination is frequently facilitated through the use of animations. For instance, educational content on topics like AIDS is commonly conveyed through animated presentations. These animations are typically shared on online video platforms such as YouTube~\cite{Youtube}. Additional curation is essential to tailor them for model training purposes. However, in certain cases, unique animations must be created from scratch, a process that can be labor-intensive and time-consuming. 

Compared to real-world videos, animations do not contain as many details or adhere to physical laws as strictly, which makes the generation of animations simpler than that of real-world videos. However, medical animations still need to follow specific rules such as the patterns of disease onset and the fundamental procedures of treatment. SOTA open-source video generation models such as Open-Sora~\cite{opensora, opensora-plan} and Movie-Gen~\cite{MovieGen} may be suitable for handling this generation task. However, although both of them possess powerful video generation capabilities, enabling them to fully comprehend medical concepts and generate medical content remains a challenging task.
. In addition to the video component, these videos also need to generate audio simultaneously. MovieFactory~\cite{zhu2023moviefactory} and Movie Gen~\cite{MovieGen} can be desired methods to complete the above tasks for their capability to generate audio suitable to the scenarios.

\subsection{Synthetic Medical Data for Training AI Models}
One significant function of video generation is the ability to enrich datasets to prevent overfitting which is a common concern when fitting on a limited scale of data, especially in the medical field because medical data is often scarce. Synthetic data could be one of these solutions as demonstrated in the recent foundation model approaches~\cite{qiu2023visionfm}. 
\begin{figure}[htb]
    \includegraphics[page=1, width=1\columnwidth]{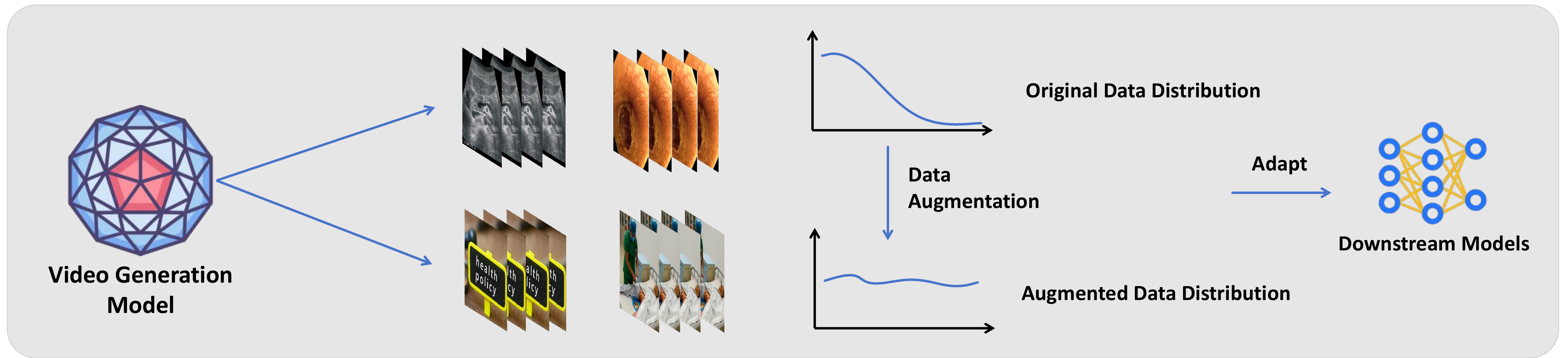}
    \caption{Video generation model synthesizes medical videos for training downstream AI models.}
    \label{fig:Architecture}
\end{figure}
\subsubsection{Data Augmentation}

Many works~\cite{frid2018gan, shin2018medical, xu2024cross} have proved that synthetic data augmentation can be an effective tool for improving a model's performance and robustness in downstream tasks such as classification and segmentation. Although recent work~\cite{wang2024generated} has proved that too much synthetic data will cause model degradation, a moderate amount of synthetic data is beneficial to model performance~\cite{liu2024can}.

Long-tailed distributions are common in medical datasets~\cite{wu2024medical}, resulting in the model being more inclined to the header data (i.e., the class with a large number of samples) during training, while the learning of the tail data (i.e., the class with a small number of samples) is relatively weak~\cite{gu2022tackling}. The generation of tail data helps handle the bias situation, making the model more sensitive to rare diseases. Elberg et al.~\cite{elberg2024longtail} proposed a method for image augmentation in long-tailed CT data by creating a modified separable latent space to mix head and tail class examples. Some other methods such as SUGAN~\cite{colleoni2024guidedgen}, CFIGGAN~\cite{CFIGGAN}, and the method proposed by Go et al.~\cite{go2024generation} can also control the generation of images based on descriptions and other constraints to achieve the purpose of data augmentation. Deo et al.~\cite{deo2024one}proposed content-space assessment to select skin lesion images generated by one-shot-gan, which calculates and maximizes the distance between generated images and original images in spatial and content domains.  


\subsection{Diagnostic Assistance}

Synthesis videos have the potential to offer doctors more comprehensive diagnostic evidence, enhancing the accuracy of diagnoses and potentially lowering diagnostic costs. This includes reducing high medical examination fees and minimizing radiation exposure from these procedures.~\cite{zhao2024largeDSA}.
\begin{figure}[htb]
    \includegraphics[page=1, width=1\columnwidth]{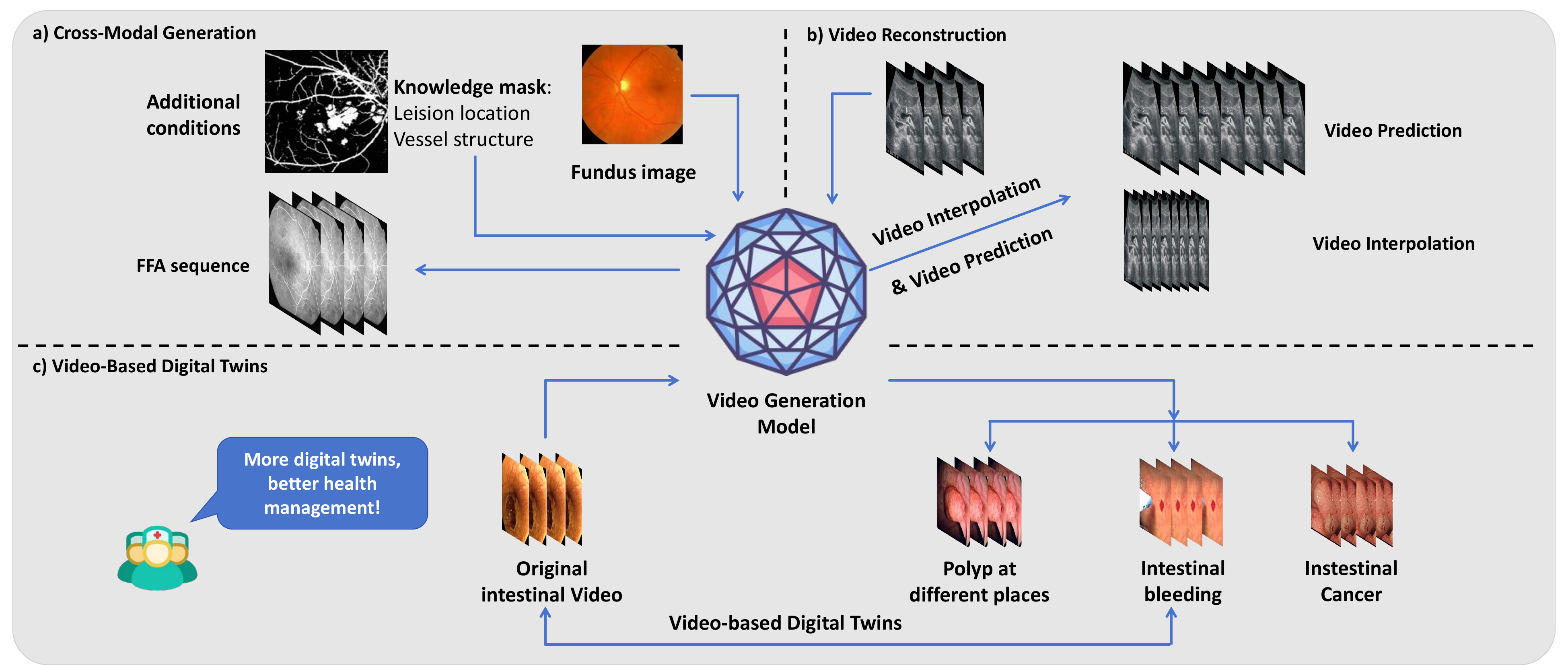}
    \caption{Video generation model for diagnostic assistance, including a) cross-modal generation; b) video reconstruction; and c) video-based digital twins.}
    \label{fig:Architecture}
\end{figure}
\subsubsection{Cross-modal Generation}\label{cross modal generation}

In disease screening, it is usually necessary to undergo multiple tests in order for the doctor to reach a final diagnosis. For example, when diagnosing eye diseases, some patients may have to take both fundus and FFA examinations. However, certain examinations are usually costly and can also cause harm to the patient, such as FFA which requires the injection of contrast agents into the patient's blood vessels and may also cause allergic reactions. Therefore, obtaining these modalities via video generation techniques can  reduce the cost of examination and minimize harm to the patient, making it an effective method of auxiliary diagnosis.

Cross-modal generation requires data pairs for training. Fundus2Video~\cite{fundus2video} has prepared 350 pairs of fundus-FFA data from 350 patients. At the same time, more modalities can be added to supplement the domain gap between the original data and the video modality. In Fundus2Video, the knowledge mask serves as a condition to guide the generation of FFA videos, and under its guidance, accurately generating lesions becomes more feasible.

\subsubsection{Digital Twins}\label{Digital Twins}
The digital twin is a virtual model designed to create a digital replica of its physical counterpart, such as a person. Medical digital twin technology can reflect patient data onto a digital entity, thereby simulating the patient's condition. Video editing, a kind of algorithm designed to alter video content, such as adjusting the stylistic elements of the video, substituting objects within the footage, and altering the video background, is an effective tool for implementing digital twins. Based on existing patient video data, doctors can obtain a copy of video data that includes lesions or abnormalities through video editing. With more diagnostic evidence, doctors can also diagnose and manage the patient's condition from a broader perspective.

\subsubsection{Medical Video Reconstruction}
One application of video reconstruction is to enhance video quality in both temporal and spatial dimensions, resulting in a more realistic and smoother viewing experience. Videos with low frame rates and missing frames can be supplemented by video interpolation and video prediction methods, respectively, to achieve diagnostic effects comparable to those of the original high-frame-rate videos or complete videos.

Video interpolation algorithms can be used in medicine to compensate for the imagery resulting from low frame rate scanning. It focuses on flow-based~\cite{niklaus2018context, liu2019deep, xu2019vqi, zhao2024largeDSA}, kernel-based~\cite{niklaus2017video, deng2019self}, and phase-based methods~\cite{meyer2015phase, meyer2018phasenet} before. LargeDSA~\cite{zhao2024largeDSA} performs video interpolation in DSA sequence and prediction through flow prediction, reducing the radiation from examinations and lowers the cost of diagnosis.

Video prediction can to some extent provide additional diagnostic evidence for doctors or be used to repair missing parts of a video. Optical flow~\cite{wu2020future, wu2022optimizing, hu2023dynamic}, depth information~\cite{liu2023meta, luc2017predicting, bei2021learning}, and motion information~\cite{jang2018video, akan2021slamp, wang2023leo} can be used to guide video prediction, and later, diffusion models~\cite{gu2023seer, zhang2024extdm} have also shown strong capabilities in the field of video prediction. 

\subsubsection{Other Application}

In addition to generating videos as direct diagnostic evidence, video generation methods can recreate the scenes described by patients, allowing doctors to gain a clearer understanding of the sequence of events, thereby assisting in diagnosis. For example, based on the patient's description of his/her injury, the video generation model can generate corresponding scenes from his/her description.

\subsection{Biomedical Simulation}
One compelling reason video generation holds significant potential in the biomedical field is its ability to dynamically simulate physiological activities, such as heartbeats. Researchers can explore various heartbeat states simply by adjusting text prompts. This straightforward approach can greatly enhance our understanding of physiology and disease mechanisms.
\begin{figure}[htb]
    \includegraphics[page=1, width=1\columnwidth]{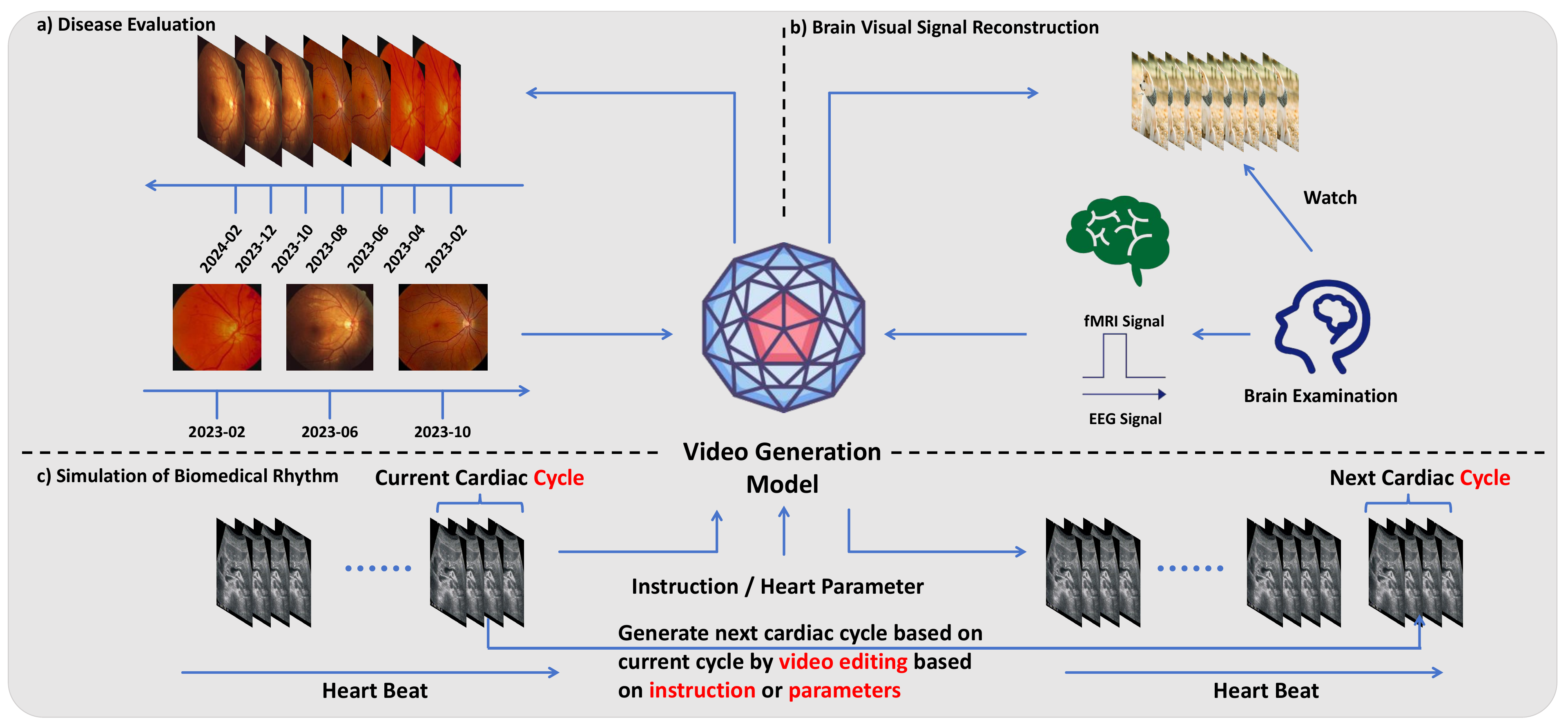}
    \caption{Video Generation model for biomedical simulation, including 1) disease evaluation; 2) brain visual signal reconstruction; and 3) biological rhythm simulation.}
    \label{fig:Architecture}
\end{figure}
\subsubsection{Simulation of Biological Rhythm}
The biological rhythm refers to the phenomenon that various functional activities within an organism repeat cyclically in a certain time, exhibiting rhythmic changes, such as the breathing of the lungs, and the beating of the heart. Scientists have built physical models to simulate the phenomenon~\cite{park2024biorobotic} that is able to simulate the structure, function, and movement of a heart. However, such artificial organs only have a shelf life of a few months at most, and maintenance and production require considerable time and economic costs. A video generator as a medical simulator can be an affordable and effective alternative to this process.

Biological rhythms are cyclical, for example, the heartbeat cycle is about $0.8$ seconds and the breathing cycle is about $4$ seconds. Existing technologies can generate a few seconds of high-quality video, which can fully cover a complete cycle. If the video generation model is capable of generating the video of the next cycle based on the video of the previous cycle according to the text prompt, the model can work as a simulator to simulate whole biological rhythms. 


\subsubsection{Simulation of Disease Evolution}
Typically, patients can only assess their condition during examinations, leaving doctors unable to ascertain the previous evolution of the disease or simulate its progression between two appointments. However, video generation models hold the promise of capturing images/videos of symptoms at various time points, allowing for the visualization of the disease's progression between these moments and enabling predictions about its future trajectory.

Predicting the disease evolution between two time points is a video interpolation problem. Based on the first and last frames, it is possible to generate the video between the two frames. But unlike traditional video interpolation tasks, which involve upsampling a low-frame-rate video to a high-frame-rate video to make the video smoother, video interpolation for analyzing disease progression is based on the interval between two medical examinations, which may be one year apart, to generate the disease's evolution process. This type of video interpolation spans a significant temporal dimension. For instance, if two CT scans are conducted a year apart and doctors require more closely spaced examination data, a generative model can be employed to produce twelve CT images to simulate the results of monthly check-ups. By employing the aforementioned methods, the progression and trajectory of diseases can be visualized, providing explainability for AI-assisted medical diagnostics, and offering significant assistance for medical prognosis and disease comprehension.

In this task, the collected data does not have to be genuine video data, but image sequences, with a potentially large time span, ranging from days to months. There are already some datasets that have compiled this kind of data~\cite{dai2024deep, li2021causal}. DeepDR plus~\cite{dai2024deep} curated a dataset containing $717308$ fundus images from $179327$ patients recording their physical examinations for a period of seven years. In order for the model to better understand the evolution of diseases, the image-time series data should be more dense.

\subsubsection{Brain Activity Reconstruction}
Brain activity reconstruction aims to reconstruct visual contents seen by an individual by analyzing the neural activity of the brain during visual perception.  Several works~\cite{sun2024contrast, shen2019deep, nemrodov2018neural, kupershmidt2022penny, wang2022reconstructing, chen2024cinematic, yeung2024neural, bauer2024movie} focus on reconstructing images from brain signals such as functional magnetic resonance imaging (fMRI) or electroencephalo-graph (EEG) to visualize the dynamic activities of brain. 

The dataset includes videos and their corresponding brain signals such as fMRI or EGG signals~\cite{wen2018neural}. It is worth noting that the temporal resolution of brain signals is much lower than the frame rate of videos~\cite{lu2024animate}, indicating that the information from brain signals may not be sufficient to fully represent video information, which can affect the reconstruction outcome.

There is a need for lightweight networks or optimization of the inference speed of generative models to enable rapid response to changes in brain signals during the inference process. Recently, some works have focused on fast image generation~\cite{zheng2024cogview3, chen2024pixart}. For example, Pixart-$\delta$ is capable to generate a $1024 \times 1024$ image on A100 GPU in $0.5$ seconds. DiffSHEG~\cite{chen2024diffsheg} can generate real-time 3D emotions and gestures guided by sound. 

Additionally, a more appropriate and efficient encoder for brain signals is needed to extract important and useful information. Brain signals should be further aligned with videos. fMRI-PTE~\cite{qian2023fmri} pretrained on over $40000$ fMRI subjects and work as a good feature extractor. To achieve better visual reconstruction effects, the signal encoder also needs to be optimized.

\section{Risks of Generated Biomedical Videos}\label{Risks}

Although there is great anticipation regarding the use of AI-generated video content for medical purposes, possible risks must also be considered. The main concerns that have been identified in the use of such applications include hallucinations~\cite{bhattacharyya2023high}, bias~\cite{ratwaniaddressing}, violation of privacy~\cite{guan2019artificial}, lack of AI accountability~\cite{duffourc2023proposed}, among others. In this section, we elaborate on each of these risks.

\subsection{AI Hallucination }
Hallucinations refer to incorrect predictions made by AI models that can be caused if the dataset on which it is trained is biased or low-quality leading to overfitting~\cite{maleki2024ai}. This has been most evidently reported in the medical field when the AI model is asked to provide diagnostic suggestions or interpret clinical data~\cite{qiu2024application}. In a recent study~\cite{bhattacharyya2023high}, GPT-3.5 was used to write medical articles with references. After cross-checking the references across multiple repositories, it was found that 47 \% of the references were fabricated and another 46 \% were authentic but inaccurately interpreted. Only 7 \% of the references were authentic and accurate.

This makes applications such as patient-doctor interface usage using AI questionable as the video content of remedies or treatments made by such models can often be misleading to the patients. This would lead to a requirement for stronger scrutiny by the doctors when using such platforms, making their tasks more tedious and prone to error. As the references used or cited by such platforms are often also false~\cite{bhattacharyya2023high}, there is a risk of spreading medical knowledge that is harmful to the treatment of patients.

Hallucinations can also negatively impact video-generated applications such as surgical training. The video generated by AI for training and suggested surgical procedures might suggest techniques or instruments that are not conventionally used by surgeons. This can lead to great misinformation and confusion among the trainees who have no prior experience of surgical procedures. Secondly, such surgical videos would have to be able to accurately depict the anatomical structures of the human body while undergoing surgery. Due to the poor grounding of the model, it is possible that the generated content might not accurately reflect, with the required details, the anatomical definition. This would further lead to major confusion in the trainees who will experience vastly different anatomical structures when operating on a patient.

\subsection{Bias in Generated Content }

AI models can often reflect systemic human biases, which can appear in various forms such as biases related to gender, race, ethnicity, age, socioeconomic status, and geographic location~\cite{khan2023drawbacks}. In healthcare, these biases typically stem from biased and imbalanced datasets, which may reflect structural discrimination embedded in data collection methods or in the ways doctors treat patients~\cite{khan2023drawbacks}.
This leads to the risk to worsen health disparities through biases in the data training the algorithm, and in the algorithm's intent.  For instance, women and minority groups are historically under-represented in cardiology~\cite{tat2020addressing}, and the bulk of current evidence-based medicine might not necessarily apply to these populations. Hence, if an AI model is trained on historical medical data that predominantly features male patients, the generated educational videos might disproportionately depict male patients undergoing cardiological procedures. This could lead to a lack of representation of female patients in these videos, potentially causing medical professionals to be less familiar with recognizing and treating heart conditions in women. Hence trained physicians are less familiar with the unique cardiological challenges of women.

\subsection{Lack of AI Accountability }

The proposed EU Directives for AI liability, namely the Product Liability Directive (PLD) and the AI Liability Directive (AILD), aim to provide uniform liability rules for AI-caused harm~\cite{duffourc2023proposed}. However, they leave significant gaps, particularly concerning black-box medical AI systems. These systems use complex and opaque reasoning~\cite{sun2023survey}, making it difficult for patients to successfully sue manufacturers or healthcare providers for injuries. For instance, serious medical harm can be done to patients being subject to incorrect medical treatment from the generated AI content.

Additionally, the directives do not fully address these potential liability gaps, leading to uncertainty for manufacturers and healthcare providers in predicting liability risks associated with these AI systems~\cite{duffourc2023proposed}. This uncertainty may hinder the development and use of beneficial black-box medical AI systems, ultimately impacting patient care and innovation in the medical field.

\subsection{Breach of Private Medical Data Laws}

The use of medical AI as of now is greatly restricted to communities with significant advancements in public healthcare~\cite{guan2019artificial}. Moreover, large number of communities do not have adequate medical databases, such as Electronic Health Records or Surgical Videos. 

Even in societies with strong records of structured digital health records, there are certain privacy laws~\cite{guan2019artificial} that need to be satisfied before gaining access to them for training AI models.  A major risk is a breach of the privacy of patients when there is a failure in maintaining anonymity~\cite{sepas2022algorithms} in the generated video content. This looks like revealing personal information or images of patients in the promotional videos being used to educate the public about certain diseases or conditions. Such errors in the system can lead to distrust among users and make them less willing to share their medical data for the purpose of generative AI in medicine.

\subsection{Model Jailbreaking and Content Safety}
Model jailbreak refers to the use of prompts to bypass the safety restrictions of generative models such as LLMs, inducing the model to produce outputs that violate its original design intent or security guidelines. These attacks exploit vulnerabilities in the model's internal processing mechanisms, potentially leading the model to generate harmful, inappropriate, unwanted, or even illegal content~\cite{wei2024jailbroken}.

In the field of medical video generation, model jailbreaking refers to the inappropriate generation of videos. For example, when a patient is consulting about the treatment options for glaucoma, but the virtual consultation model generates a video of glaucoma surgery, which is graphic and inappropriate. Alternatively, the video may display private information such as the patient's age and name.
Such content is not suitable and may cause discomfort to the patient. The main reason may be that due to the diversity of the patient's input, it bypasses the safety mechanisms of the generation model, constructing keywords and phrases that can activate specific features or content~\cite{yang2024sneakyprompt}. Adversarial training~\cite{li2024one, li2024aroid} should be considered in building safer video generative models.

\subsection{Explainability and Reliability}

Video generation models, acting as black boxes, obscure the process by which input conditions are mapped to the generated videos. For example, the video generation model proposed by Nguyen Van Phi et al.~\cite {van2023echocardiography} attempts to generate ultrasound videos of cardiac motion by incorporating semantic masks of heart structures. Although the generated results can map the structures to the videos with a relatively high degree of accuracy, it remains uncertain whether the structure of the generated video perfectly aligns with the semantic mask, and whether the generated video corresponds to the actual conditions of cardiac ultrasounds, including whether aspects such as vascular quantification metrics, cardiac contractility indicators, and vascular morphological descriptions in the generated videos are consistent with real situations, which are difficult to substantiate.

Due to the inability to explain what diagnostic medical information is contained within synthesized medical videos, using them as a basis for diagnosis poses significant risks. Enhancing the transparency of the model so that medical professionals can comprehend the reasoning behind its predictions is crucial for the widespread application of medical video generation models.

\section{Conclusions}\label{Conclusion}

This article discusses the development of video generation techniques, with a focus on biomedical video generation. It provides a comprehensive overview of the techniques and datasets available for biomedical video generation, existing and potential usage scenarios, as well as metrics for evaluating the quality and accuracy of generated videos. We hope that this article can inspire related research fields to find novel solutions, thereby deepening and broadening the development and applicability of video generation in biomedicine.

\printbibliography

\end{sloppypar}
\end{document}